\documentclass[journal]{IEEEtran}

\usepackage{graphicx}
\usepackage{amsmath,amsfonts,amssymb,graphicx}    
\usepackage{subfigure}
\usepackage{indentfirst}
\usepackage{listings}
\usepackage{cite}
\usepackage{xcolor}
\usepackage{multirow}
\usepackage{bm}
\usepackage{array}
\usepackage{amsthm}
\usepackage{colortbl}
\usepackage{siunitx}
\usepackage{gensymb}
\usepackage{booktabs}
\usepackage{stfloats}
\usepackage[colorlinks=true,
            linkcolor=black, 
            anchorcolor=black, 
            citecolor=black, 
            urlcolor=blue
            ]{hyperref}
\usepackage[final]{changes}
\usepackage{booktabs} 
\usepackage{makecell}
\usepackage{bbding}
\UseRawInputEncoding
\usepackage{epstopdf}
\usepackage{threeparttable}
\usepackage{orcidlink}

\usepackage[linesnumbered,ruled,vlined]{algorithm2e}

\begin{document}

\title{URPlanner: A Universal Paradigm For Collision-Free Robotic Motion Planning Based on Deep Reinforcement Learning}

\author{Fengkang~Ying\orcidlink{0000-0003-1931-378X},~\IEEEmembership{Graduate Student Member,~IEEE}, Hanwen~Zhang\orcidlink{0009-0007-1785-1732}, Haozhe~Wang\orcidlink{0000-0001-9126-2425},~\IEEEmembership{Graduate Student Member,~IEEE}, Huishi~Huang\orcidlink{0000-0002-0533-7847}, ~\IEEEmembership{Graduate Student Member,~IEEE}, Marcelo~H.~Ang~Jr.\orcidlink{0000-0001-8277-6408},~\IEEEmembership{Senior Member,~IEEE}

\thanks{
This research is supported by the National Research Foundation (NRF), Prime Minister’s Office, Singapore under its Campus for Research Excellence and Technological Enterprise (CREATE) programme. The Mens, Manus, and Machina (M3S) is an interdisciplinary research group (IRG) of the Singapore MIT Alliance for Research and Technology (SMART) centre. \textit{(Corresponding author: Haozhe Wang.)}}

\thanks{F. Ying and H. Wang are with the Integrative Sciences and Engineering Programme, NUS Graduate School, National University of Singapore, 119077, Singapore, and Advanced Robotics Centre, National University of Singapore, 117608, Singapore. (e-mail: fengkang@u.nus.edu; wang\_haozhe@u.nus.edu).}

\thanks{H. Zhang, H. Huang, and M. H. Ang Jr. are with the Department of Mechanical Engineering, College of Design and Engineering, National University of Singapore, 117575, Singapore and Advanced Robotics Centre, National University of Singapore, 117608, Singapore. (e-mail: hanwen.zhang@u.nus.edu; huishi.huang@u.nus.edu; mpeangh@nus.edu.sg).}
}

\markboth{}%
{Shell \MakeLowercase{\textit{et al.}}: Bare Demo of IEEEtran.cls for Journals}
\maketitle

\begin{abstract}
Collision-free motion planning for redundant robot manipulators in complex environments is yet to be explored. Although recent advancements at the intersection of deep reinforcement learning (DRL) and robotics have highlighted its potential to handle versatile robotic tasks, current DRL-based collision-free motion planners for manipulators are highly costly, hindering their deployment and application. This is due to an overreliance on the minimum distance between the manipulator and obstacles, inadequate exploration and decision-making by DRL, and inefficient data acquisition and utilization. In this article, we propose URPlanner, a universal paradigm for collision-free robotic motion planning based on DRL. URPlanner offers several advantages over existing approaches: it is platform-agnostic, cost-effective in both training and deployment, and applicable to arbitrary manipulators without solving inverse kinematics. To achieve this, we first develop a parameterized task space and a universal obstacle avoidance reward that is independent of minimum distance. Second, we introduce an augmented policy exploration and evaluation algorithm that can be applied to various DRL algorithms to enhance their performance. Third, we propose an expert data diffusion strategy for efficient policy learning, which can produce a large-scale trajectory dataset from only a few expert demonstrations. Finally, the superiority of the proposed methods is comprehensively verified through experiments. 
\end{abstract}

\begin{IEEEkeywords}
Collision-free motion planning, parameterized task space, universal obstacle avoidance reward, augmented policy exploration and evaluation, expert data diffusion.
\end{IEEEkeywords}

\IEEEpeerreviewmaketitle

\section{Introduction}
\subsection{Motivation and Related Work}
\IEEEPARstart{I}{N} recent years, robot manipulators have been increasingly utilized across a diverse array of environments and scenarios, ranging from industrial manufacturing and processing to daily service and recreation \cite{Lv2023}. Particularly, redundant manipulators can move more flexibly and are more adaptable to the growing complexity of the environment they operate in due to their enhanced ability to avoid collision and singularity \cite{Safeea2021}. However, the increased environmental complexity and robotic redundancy have also significantly complicated the collision-free motion planning problem. 

Traditional motion planning (TMP) methods, such as rapidly exploring random trees (RRT) and probabilistic roadmap (PRM), are probabilistically complete \cite{Qureshi2021}. This property ensures that a solution can always be found if it exists, as long as the search time approaches infinite. However, the path they find is non-optimal \cite{Ruan2022}. Their improved variants RRT$^*$ and PRM$^*$ can find asymptotically optimal paths, but the optimization ability is limited as they do not leverage information from past experiences \cite{XiangjianTII}. In comparison, learning-based planning methods can learn from previously obtained samples to improve planning and optimization efficiency.

Besides the optimization problem, these TMP methods also exhibit several deficiencies in other aspects. For example, they often fall short of efficiently handling the high-dimensional space, causing the degradation of performance as the complexity of the environment increases \cite{Faust2018}. However, learning-based methods can perceive and represent high-dimensional environmental states by neural networks, making the planning more tractable in complex environments. Moreover, for TMP methods, each planning task requires repeatedly searching the whole environment from scratch. In contrast, learning-based methods can be trained to directly infer an optimal solution without going through the search process. 

Additionally, for the motion planning of manipulators, these traditional methods rely on solving inverse kinematics (IK) to map the trajectories from Cartesian space to joint space. In this process, extra considerations regarding the collision avoidance of the entire manipulator are required \cite{Fengkang2024}. For manipulators with less than six degrees of freedom (DoFs), TMP methods usually go through all possible IK configurations to find the optimal one. However, the above procedures can be quite complicated and time-consuming for redundant manipulators since there are infinite IK solutions for each waypoint of the trajectory. By contrast, some researchers have designed learning-based methods that can bypass solving the IK and directly learn a policy capable of reaching the target point by constantly controlling the joint rotation of manipulators. Such methods exhibit better universality as they can be applied to manipulators with arbitrary DoFs without designing a customized IK solving and planning method \cite{FengkangTII, FengkangASME, Fengkang2024}.

Recently, the intersection of DRL and robotics has piqued much attention since it enables robots to autonomously learn optimal policies for versatile tasks. Such policies can adapt to diverse and unpredictable environments and can continuously improve performance through self-learning and adaptation \cite{Safaoui2024,R11}. Thus, DRL is particularly suited for addressing the intricate robotic motion planning problem that is extremely challenging for traditional methods. For example, regarding trajectory generation tasks, Ying \textit{et al.} \cite{FengkangTII} proposed an extensively explored and evaluated actor-critic (E3AC) algorithm to realize an IK-free motion planning scheme, meaning that the method is independent of solving the IK and can be applied to manipulators with any DoFs. In \cite{FengkangASME}, the authors proposed a multi-policy agent that could effectively handle sequential trajectory planning for multi-process robotic tasks. In terms of motion planning for pick-and-place tasks, Hu \textit{et al.} \cite{Hu2023} leveraged the proximal policy optimization algorithm and inverse reinforcement learning techniques to plan the motion for grasping living objects such as fish. The deep Q-network algorithm has been widely used to plan the motion sequence of grasping cluttered objects \cite{Zeng2018, Deng2019, Berscheid2019}. For compliant motion planning problem, Hou \textit{et al.} has adopted deep deterministic policy gradient (DDPG), twin delayed DDPG (TD3), and soft actor-critic (SAC) algorithms to implement multiple peg-in-hole assembly \cite{Hou2022, Hou2023}. Xu \textit{et al.} \cite{Xu2022Piano} applied the SAC algorithm to enable a robot hand to learn the techniques of playing the piano. For soft robot arms and deformable manipulation, DRL has been leveraged to train a pneumatic-actuated soft robot to throw objects \cite{Bianchi2023} and to manipulate deformable ropes \cite{Pecyna2022}.

Despite the advancements in DRL-based motion planning for manipulators, three main deficiencies persistently hinder its effectiveness and efficiency in real-world deployment.

Firstly, the obstacle avoidance ability of the manipulators is rarely considered in the aforementioned studies and remains relatively unexplored \cite{R11}. Regarding this, the design of appropriate and task-specific reward functions is crucial, as it directly affects the learning outcomes of DRL policies \cite{Qureshi2021}. One approach is to set the reward as zero when a collision is detected \cite{XiangjianTII, CYY2023}. However, such sparse rewards result in many different samples with the same zero feedback, significantly degrading the policy learning process \cite{Fengkang2024}. As for dense rewards, current methods for obstacle avoidance heavily rely on the minimum distance information between the robot and obstacles. For example, Sui \textit{et al.} \cite{Sui2021} designed a linear reward that is proportional to the minimum distance. In comparison, Ying \textit{et al.} \cite{Fengkang2024} proposed a quadratic reward function based on the minimum distance, which can better balance obstacle avoidance against action exploration. However, these methods idealistically assume that the minimum distance can be accurately and efficiently obtained, while in practice, it requires the geometry of all the objects, sufficient sensors, and complex analyses. 

Secondly, current DRL methods exhibit limitations in their exploration and decision-making capabilities, leading to inefficient and unstable training performance \cite{Fengkang2024}. Regarding the exploration, most DRL algorithms, such as DDPG and TD3, produce only a single action based on the current state and then explore only once by adding a random noise. However, the single-action approach in exploration causes a lack of diversity and scalability, and the randomness may even worsen the performance of the original action. To enhance exploration, previous studies have introduced a multi-actor structure to generate multiple action candidates \cite{Fengkang2024,FuXiaokuan}. However, it also made the process more inefficient. Similarly, regarding decision-making, there might be prejudice when the values of actions are estimated by only a single critic. To address this issue, an extensive evaluation architecture has been proposed, where multiple critics are integrated to decide the optimal action in a more unbiased way \cite{Wu2018,FuXiaokuan}. Nevertheless, the critic networks typically exhibit poor performance during early training stages, leading to unsatisfactory value evaluation. 

Thirdly, data acquisition and utilization pose significant challenges to deploying DRL algorithms. While DRL does not necessitate labeled datasets, policy optimization still relies on large-scale data obtained by random interaction without specific directions. Thus, it is difficult for DRL to optimize the current policy efficiently unless ample high-quality data have been obtained \cite{Fengkang2024}. The phenomenon is more severe in the early training stage, as the networks have not been sufficiently trained. Regarding the problem, prioritized experience replay has been leveraged to make full use of the data with relatively higher quality \cite{Ma2021, CHEN202264}. However, as a technique for DRL with sparse reward, its effect can be limited if the reward is dense \cite{FengkangTII}. Ying \textit{et al.} \cite{FengkangASME} improved the learning efficiency by sampling data with high reward in priority, but ranking large-scale data greatly increases the computational costs. Previous studies have also used expert demonstrations to provide high-quality data, which are beneficial to the training of DRL \cite{FengkangTII, Ma2021}. Nevertheless, acquiring substantial expert data requires extensive manual effort, while a small number of demonstrations may have a very limited effect on training DRL policies.


\subsection{Contributions and Organizations of the Article}
In response to the above discussions, this article proposes a universal paradigm for collision-free robotic motion planning based on DRL, named URPlanner. URPlanner has several intriguing advantages over existing DRL-based methods. First, we use parameterized space to represent the environment, which makes URPlanner platform-agnostic and highly cost-effective in both training and deployment. Second, the trained policy can be directly applied to robots in simulators or the real world without requiring fine-tuning. This is possible because both policy training and trajectory generation use the real robot's kinematic parameters in the parameterized space.

The definition of some critical terms in this article is given as follows:

1) \textbf{platform-agnostic:} in our manuscript, we assume static obstacle layouts, and the term \textit{platform-agnostic} refers specifically to two aspects: (i) the training of our motion planning scheme is independent of interaction with any specific simulated or real-world robotic platform, owing to the proposed parameterized task space; and (ii) the motion planning policy trained in the parameterized task space can be applied to the corresponding task scenario built on any simulated or real-world platform.

2) \textbf{universal:} the term \textit{universal} means that the proposed URPlanner can, in theory, be applied to train any manipulator in any given task scenario. However, the Denavit-Hartenberg parameters of the manipulator and the radii of the cylinders encapsulating the links should be provided.

The main contributions of this article are:

1) A universal obstacle avoidance reward (UOAR) is designed. The reward is dense, independent of minimum distance, and can be easily calculated. To achieve this, we first build a parameterized environment where the obstacles and manipulator are represented by the composition of bounding boxes and line segments, respectively. Then, the overlap of each line segment with each bounding box is inferred as the UOAR. Finally, it is combined with a pose reward to solve the general robotic motion planning problem based on DRL.

2) An augmented policy exploration and evaluation (APE2) algorithm is proposed. APE2 can be applied to various DRL algorithms, reinforcing their exploration and decision-making capabilities. It generates diverse action candidates under every state by enhanced action exploration. Then, a hybrid policy evaluation approach considering both immediate and long-term environmental feedback is proposed, which provides a more accurate and comprehensive evaluation throughout the training process.

3) An expert data diffusion (ED2) strategy is proposed, where abundant expert data are autonomously generated to facilitate the policy learning of DRL. First, a diffusion model is exploited to produce a large-scale trajectory dataset from only a few demonstrations, minimizing the need for manual participation. Then, two different methods for integrating the dataset with DRL are explored and compared: one creates a synergy between behavior cloning (BC) and DRL, while the other directly utilizes the dataset to supply additional high-quality samples for policy optimization.


The rest of this article is organized as follows: Section II formulates the motion planning problem. Section III describes the details of environment parameterization and the proposed UOAR. Section IV introduces the proposed APE2 algorithm. Section V presents the design of ED2. Section VI summarizes the overall implementation of URPlanner. In Section VII, various experiments are conducted and discussed. Section VIII presents further discussions on the universality of URPlanner, its generalization across different scenarios, the strategies for providing demonstrations to ED2, and some extensions of URPlanner. Finally, some conclusions are given in Section XI.

\section{Problem Formulation}
In this section, we present a general definition of collision-free motion planning problems based on DRL.

For manipulators, a motion planning problem is to find a sequence of collision-free robot configurations represented as waypoints and the corresponding joint angles \cite{xie2023}. The problem can be formulated as a Markov decision process. At time step $t$, the agent captures the state $\bm{s}_t$ (such as the target pose) of the environment and takes an action $\bm{a}_t$ (such as the joint rotations). The interaction transits the state to $\bm{s}_{t+1}$, and a reward $r_t$ is obtained as the environmental feedback. 

By defining the motion planning of manipulators as a Markov decision process, DRL algorithms can be employed to learn an optimal policy $\pi^*$ by maximizing the future cumulative reward $G_t$
\begin{align}
G_t &= r_t + {\xi}{r_{t+1}} + {\xi^2}{r_{t+\rm 2}} + {\cdots} + {\xi^{n-t}}{r_n} \nonumber \\
&= r_t + {\xi}{G_{t+1}}    \label{eq01}
\end{align}
where $\xi$ is a discount factor. $G_t$, also known as the long-term return, represents the cumulative reward an agent expects to receive over time, starting from the current state and following a given policy. In DRL, the Q-value produced by the critic network serves as an estimate of the long-term return.

In comparison to \cite{Fengkang2024}, we remove the minimum distance information from $\bm{s}_t$ to make it more general and practical
\begin{align}
{\bm s}_t = [{\bm q}_t, {\bm p}_{\rm T},{\bm p}_{\rm G}, {\Delta \rm p}, {\Delta \rm o}, D] \label{eq02}
\end{align}
${\bm q}_t$ is the configuration (i.e. joint angles) of the manipulator. ${\bm p}_{\rm T}$ and ${\bm p}_{\rm G}$ are the pose of the manipulator's tool center point and the goal, respectively. $\Delta {\rm p}=[\Delta x,\Delta y,\Delta z]$ and $\Delta {\rm o}=[\Delta\alpha,\Delta\beta,\Delta\gamma]$ are the position and orientation deviation between the tool center point and the goal, respectively. $D$ is a Boolean variable that equals one when the pose of tool center point is within the allowable errors.


Unlike previous studies that learn the translation and rotation of a manipulator's tool center point for motion planning \cite{Hou2021, Hou2022, Hou2023}, we let the manipulator directly learn sequential rotations of its joints to bypass IK. Hence, ${\bm a}_t$ is defined as
\begin{align}
{\bm a}_t = [{\Delta{q}} _{1},{\Delta{q}} _{2},\cdots, {\Delta{q}} _i]  \label{eq03}
\end{align}
where ${\Delta{q}} _i$ is the angular increment of joint $i$.

\section{Space Parameterization and Reward Shaping}
In this section, the task space for robotic motion planning is parameterized, with obstacles and robot manipulators described by mathematical models. A UOAR independent of minimum distance information is designed and combined with pose rewards based on the parameterized space.

\subsection{Task Space Parameterization}
\label{Sec:III-B}

\begin{figure}[!t]
\centering
\includegraphics[width=3.2in]{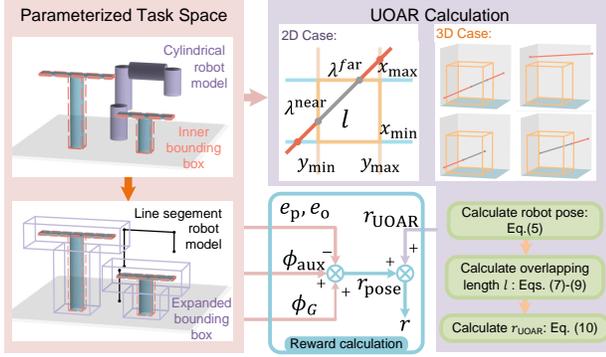}
\caption{Schematic diagram of the parameterized space and reward calculation. 
The manipulator is represented as line segments encapsulated by cylinders, while the obstacles are modeled as expanded bounding boxes, incorporating the cylinder's radius along with a safety offset. UOAR is designed based on the overlap length between the line segments and bounding boxes. This reward encourages the DRL agent to learn collision-free motion planning by minimizing the overlap length, thereby ensuring safe and efficient trajectories.}
\label{Fig2.1}
\end{figure}

As shown in Fig. \ref{Fig2.1}, the spatial occupancy of an obstacle with arbitrary geometry can be approximated using one or more axis-aligned bounding boxes (AABBs) \cite{CHEN202264}. For irregularly shaped obstacles, we recommend using multiple boxes to better approximate their spatial occupancy and minimize unnecessary reserved space. The space enclosed by AABB can be expressed by a point set defined as $Box =\{(x,y,z)\;|\;x\in [x_{\rm min},x_{\rm max}],y \in [y_{\rm min},y_{\rm max}], z \in [z_{\rm min},z_{\rm max}] \}$.


For a given manipulator, define $O_i$ as the coordinate origin of joint $i$. The geometric relationship between $O_i$ and $O_{i-1}$ can be represented by Denavit-Hartenberg (DH) parameters: link length $a_i$, link twist $\alpha_i$, link offset $d_i$, and joint angle $q_i$. Based on the DH parameters, the homogeneous transformation matrix between two adjacent joints is defined as
\begin{align}
T_i^{i-1}={\rm{Rot}}(\alpha_i){\rm{Trans}}(a_i){\rm{Rot}}(q_i){\rm{Trans}}(d_i)\label{eq3.1}
\end{align}

Further, defining the robot base coordinate as the world coordinate system, the pose of joint $i$ can be computed by 
\begin{align}
T_{i}^{0}=T_1^{0}T_2^{1}{\cdots}T_i^{i-1}=\begin{bmatrix} {{\bm R}_{i}^{0}} & {{\bm P}_{i}^{0}} \\ {\bm 0} & 1 \end{bmatrix}_{4\times4} \label{eq3.2}
\end{align}
where ${\bm R}_{i}^{0}$ and ${\bm P}_{i}^{0}=[x_{i}^{0},y_{i}^{0},z_{i}^{0}]^{\rm T}$ describe the orientation and position of joint $i$ in the world coordinate system, respectively.

Defining ${\bm{p}}_{i}(x_{i}^{0},y_{i}^{0},z_{i}^{0})$ as the coordinates of joint $i$ in the world coordinate system, each link of the manipulator can be simplified and parameterized as a line segment ${\bm s}_i$ enclosed by a cylinder with a radius $a_{\rm r}$, denoted as
\begin{align}
{\bm s}_{i}^{i-1} = {\bm{p}}_{i-1}+\overrightarrow{{\bm{p}}_{i-1}{\bm{p}}_{i}}\cdot \lambda, \;\; 0\le \lambda \le 1
\label{eq3.3}
\end{align}
The parameterized model of any given manipulator can be easily obtained according to its DH parameters. The parameter can be manually modified if necessary so that the links can be evenly encapsulated by cylinders. Additionally, some adjacent link pairs (such as links 1 and 2, links 3 and 4, links 5 and 6, and links 7 and 8 of the KUKA LBR iiwa robot or Franka Emika Panda robot) can be consolidated into a single link.

Finally, as shown in Fig. \ref{Fig2.1}, by expanding the bounding boxes to include the radius $a_{\rm r}$ of the simplified robot model and a safety offset $a_{\rm o}$, the whole collision-free motion planning problem for manipulators can be transformed into ensuring that the line segments avoid intersections with the expanded bounding boxes during motion. Here, $a_{\rm o}$ establishes a buffer zone between the manipulator and surrounding obstacles. This approach aligns with the principles of clearance-based motion planning, which aims not only to prevent collisions but also to maintain a safe distance from obstacles throughout the motion.

\emph{Remark 1: }{The parameterized task space offers significant advantages for DRL-based motion planning. First, it enables efficient calculation of overlaps between the manipulator and obstacles, facilitating the design of the proposed UOAR, which is independent of the minimum distance. Second, it allows any required information for motion planning to be analytically computed. This feature makes URPlanner platform-agnostic and highly cost-effective in both training and deployment compared to existing methods \cite{Fengkang2024,CHEN202264,XiangjianTII,FengkangTII,FengkangASME}.}

\subsection{Universal Obstacle Avoidance Reward (UOAR)}
Given a parameterized link ${\bm s}$ defined by Eq. \eqref{eq3.3}, its starting point and ending point in the world coordinate system can be determined using the transformation matrix $T_{i}^{0}$ from Eq. \eqref{eq3.2}. Let $T_{0}^{\rm{Box}}$ denote the homogeneous transformation matrix that maps the world coordinate system to the bounding box's coordinate system. Using $T_{0}^{\rm{Box}}$ and $T_{i}^{0}$, the coordinates of the starting and ending points in the bounding box's coordinate system, represented as ${\bm{p}}_{\rm s}(x_{\rm s},y_{\rm s},z_{\rm s})$ and ${\bm{p}}_{\rm e}(x_{\rm e},y_{\rm e},z_{\rm e})$, respectively, can be obtained by multiplying $T_{0}^{\rm{Box}}$ with $T_{i}^{0}$.

For a bounding box with six planes, the intersections of the line (on which the segment lies) with the two planes perpendicular to the $x$ axis can be represented as ${\bm{p}}_{\rm s}+\overrightarrow{{\bm{p}}_{\rm s}{\bm{p}}_{\rm e}}\cdot\lambda$ with the corresponding values of $\lambda$. Representing the two planes as $x\!=\!x_{\rm{min}}$ and $x\!=\!x_{\rm{max}}$, $\lambda$ can be solved by

\begin{equation}
\left\{
   \begin{aligned}
   \lambda_x^{{\rm {near}}}={\rm min}(\frac{x_{\rm {min}}-x_{\rm s}}{x_{\rm e}-x_{\rm s}},\frac{x_{\rm{max}}-x_{\rm s}}{x_{\rm e}-x_{\rm s}})\\
    \lambda_x^{\rm {far}}={\rm max}(\frac{x_{\rm{min}}-x_{\rm s}}{x_{\rm e}-x_{\rm s}},\frac{x_{\rm{max}}-x_{\rm s}}{x_{\rm e}-x_{\rm s}})\\    
   \end{aligned}
\right.
\label{eq3.5}
\end{equation}
The intersections with the planes perpendicular to $y$ and $z$ axes can be solved using the same logic, respectively.

If the line segment intersects with the given bounding box, the following condition must be satisfied
\begin{align}
{\rm max}(\lambda_x^{\rm {near}},\lambda_y^{\rm{near}},\lambda_z^{\rm {near}},0)\le {\rm min}(\lambda_x^{\rm {far}},\lambda_y^{\rm{far}},\lambda_z^{\rm {far}},1)
\label{eq3.6}
\end{align}

The method described by Eqs. \eqref{eq3.5} and \eqref{eq3.6} has been broadly adopted to detect collisions in previous studies, such as \cite{Mu2014}. However, based on the collision information, only a sparse reward function can be designed, such as assigning a reward of zero when a collision is detected \cite{CYY2023,XiangjianTII}. Such sparse rewards often hinder the efficiency of the policy learning process. In cases where no collisions are detected, prior studies like \cite{CHEN202264} and \cite{Yang2022} have proposed calculating the minimum distance between the obstacle and the robot to design a dense reward. This approach provides more informative guidance, but is often computationally expensive. In this work, we demonstrate that the overlap length between line segments and expanded bounding boxes can be leveraged to design a dense reward that effectively guides the learning of obstacle avoidance policies while maintaining computational efficiency.

When the link $\bm s$ intersects with the bounding box, $\lambda^{\rm{near}}={\rm max}(\lambda_x^{\rm{near}},\lambda_y^{\rm{near}},\lambda_z^{\rm{near}},0)$ and $\lambda^{\rm{far}}={\rm min}(\lambda_x^{\rm{far}},\lambda_y^{\rm{far}},\lambda_z^{\rm{far}},1)$. Then, the length of the overlapping part of the link is 
\begin{align}
l=\lvert \lambda^{\rm{far}}-\lambda^{\rm{near}}\rvert\cdot \big\lvert\overrightarrow{{\bm{p}}_{\rm s}{\bm{p}}_{\rm e}} \big\rvert
\label{eq3.7}
\end{align}

For a parameterized task space containing $\rm G$ bounding boxes (representing the obstacles) and $\rm J$ line segments (representing the manipulator), let $l_{jg}$ represent the overlapping length between line segment $j$ and bounding box $g$, and $L_j$ represent the length of line segment $j$. Then, the proposed UOAR is defined as
\begin{align}
r_{\rm{UOAR}}=-\frac{ \sum\nolimits_{j=1}^{\rm J}\sum\nolimits_{g=1}^{\rm G}l_{jg} 
 }  {  {\sum\nolimits_{j=1}^{\rm J}L_j}   }
\label{eq3.8}
\end{align}

\emph{Remark 2: }{Compared with existing DRL-based collision-free motion planning methods, the proposed UOAR transforms the complex problem of calculating minimum distances into a simpler problem of avoiding overlaps with the bounding boxes. This makes the training process highly cost-effective and enhances the generalization capability across obstacles with various geometries.}

To calculate the pose reward, the tool center point's pose (${\bm R}_{\rm T}$ and ${\bm P}_{\rm T}$) can be inferred by Eq. \eqref{eq3.2}. ${\bm R}_{\rm T}$ is converted to the corresponding Euler angles with respect to the goal, denoted as ($\alpha_{\rm T}^{\rm G},\beta_{\rm T}^{\rm G},\gamma_{\rm T}^{\rm G}$). Define $e_{\rm p}=\big\lvert\overrightarrow{{\bm{p}}_{\rm T}{\bm{p}}_{\rm G}}\big\lvert$ and $e_{\rm o}=|\alpha_{\rm T}^{\rm G}|+|\beta_{\rm T}^{\rm G}|+|\gamma_{\rm T}^{\rm G}|$ as the position error and orientation error, respectively. Note that both errors are normalized. Then, the pose reward is defined as
\begin{align}
r_{\rm {pose}}=-(e_{\rm p}+e_{\rm o})+\phi_{\rm {aux}}+\phi_{\rm G}
\label{eq3.9}
\end{align}
where $\phi_{\rm G}$ equals one if $e_{\rm p}$ and $e_{\rm o}$ are within allowable ranges, indicating the tool center point has reached the goal. $\phi_{\rm {aux}}$ is an auxiliary reward designed to further reduce the pose error. The setup of $\phi_{\rm {aux}}$ follows the equations (12) and (13) in \cite{FengkangASME}.

Finally, the integrative reward considering both obstacle avoidance and pose control can be defined as 
\begin{align}
r = r_{\rm{pose}} + \zeta r_{\rm{UOAR}} 
\label{eq3.10}
\end{align}
where $\zeta$ is a weight coefficient.

\section{Augmented Policy Exploration and Evaluation (APE2) Algorithm}
In this section, the detailed implementation of the proposed APE2 algorithm is introduced.

\subsection{Enhanced Action Exploration}

\begin{figure}[!t]
\centering
\includegraphics[width=3.2in]{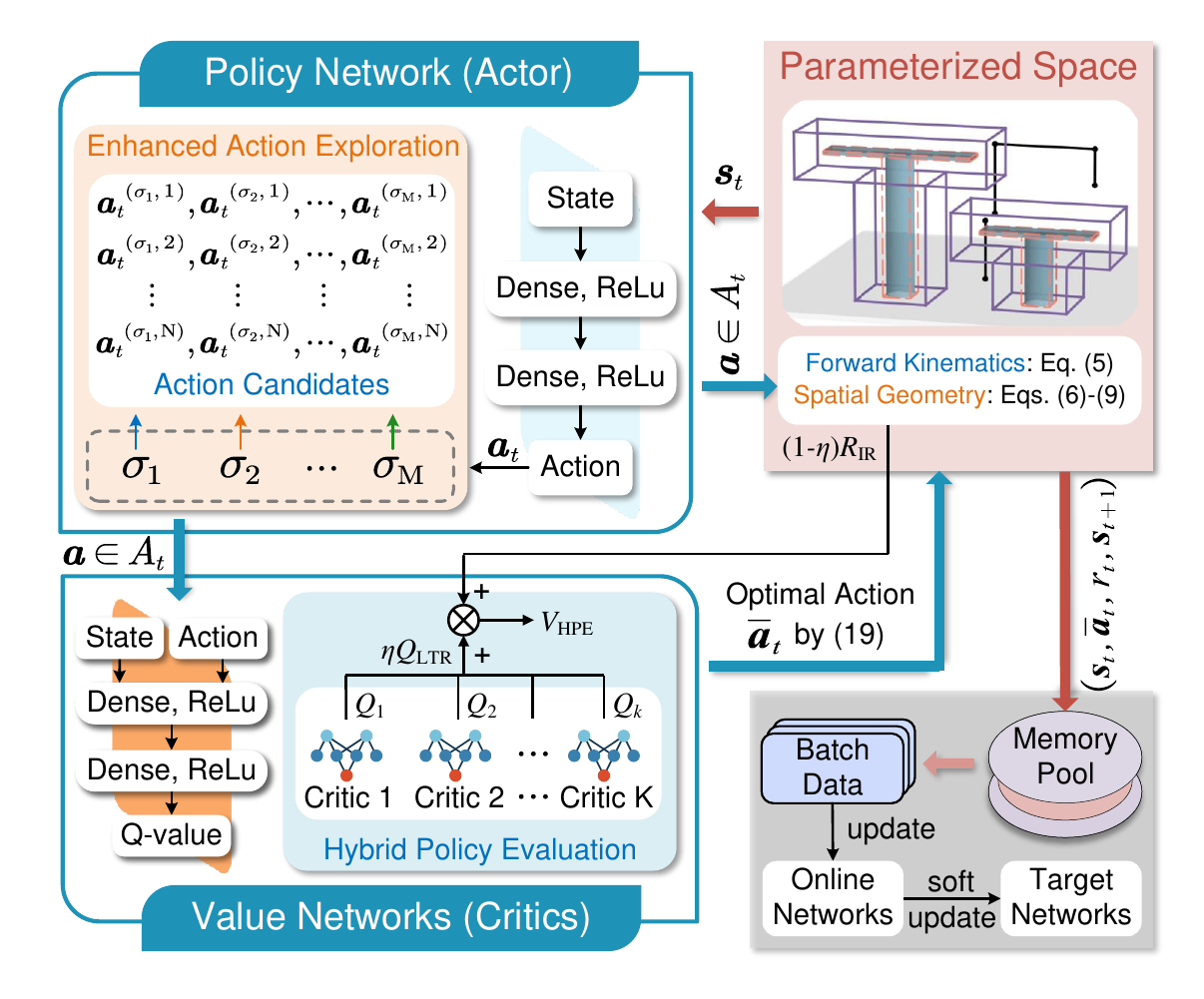}
\caption{Framework of the APE2 algorithm. For each state, APE2 efficiently generates a large pool of action candidates through enhanced action exploration. During hybrid policy evaluation, the combination of the average Q-value from multiple critics, $Q_{\rm LTR}$, and the immediate return, $R_{\rm IR}$, ensures a more accurate and comprehensive assessment throughout the training process. The agent then executes the action candidate with the highest evaluated value.}  
\label{Fig3.1}
\end{figure}

In DRL, a mapping from ${\bm s}_t$ to ${\bm a}_t$ is created by the online actor $\mu$ with network parameter $\theta^\mu$. Algorithms based on deterministic policy gradient, such as DDPG and TD3, produce only a single action ${\bm a}_t$ given the current state ${\bm s}_t$
\begin{align}
{\bm a}_t = \mu({\bm s}_t;\theta^\mu)  \label{eq4.1}
\end{align}

As shown in Fig. \ref{Fig3.1}, to improve policy exploration, at each state, the APE2 algorithm adopts an enhanced action exploration strategy to efficiently generate an action set $A_t=\{{\bm a}_t, {{\bm a}_t}^{(\sigma_m,n)} \lvert 1\le m \le {\rm M}; 1\le n \le {\rm N}; m,n \in\mathbb{Z} \}$ with diverse and large-scale action candidates
\begin{align}
{ {\bm a}_t }^{(\sigma_m,n)} = {\bm a}_t + \mathcal{N}(0,\sigma_m^2)
\label{eq4.2}
\end{align}
where $\sigma^2$ is the Gaussian variance, $\rm M$ represents the number of different Gaussian noises with diverse scales of variance. $\rm N$ means the number of times the original action ${\bm a}_t$ has been explored under each noise.

\emph{Remark 3: }{Deterministic policy gradient-based algorithms can effectively reduce the sample scale compared with stochastic policy-based ones \cite{KIM2020, FengkangTII}. However, they show limited capacity for exploring the policy. To improve this, algorithms like DDPG add a random noise to the original action ${\bm a}_t$. However, it does not increase the quantity of available actions, and the randomness may even worsen the performance of the original action. In comparison, by performing enhanced action exploration, the proposed APE2 algorithm can efficiently explore a total number of ${\rm M} \times {\rm N}+1$ action candidates for the agent to select.}

\subsection{Hybrid Policy Evaluation}
Since the enhanced action exploration provides multiple actions for APE2, it is essential to evaluate and select the optimal one for the agent to execute. The hybrid policy evaluation strategy in the proposed APE2 algorithm consists of both the Q-value, which emphasizes long-term return and the reward, which emphasizes immediate return.


As shown in Fig. \ref{Fig3.1}, the proposed hybrid policy evaluation adopts multiple critics for policy evaluation, and all the estimated Q-values are equally considered. Assuming $\rm K$ is the number of critics, the integrative Q-value $Q_{\rm{LTR}}$ is defined as
\begin{align}
Q_{\rm{LTR}} = \frac{1}{\rm K}\sum\nolimits^{\rm K}_{k=1}Q_{k}({\bm s}_t,{\bm a}_t;\theta^Q_k) \label{eq4.3}
\end{align}
where $\theta^Q_k$ ($k=1,2,\cdots,{\rm K}$) is the parameter of each online critic network $Q_k$. In practice, $\rm K$ can be any positive integer. The impact of different values of $\rm K$ on APE2 will be analyzed in the experiments.

\emph{Remark 4: }{Compared with DDPG, which has only one critic, TD3 and SAC calculate two Q-values by two critics but use only the minimum one for policy evaluation and optimization. Although this approach can prevent the overestimation problem, it assumes that the lower Q-value is always more accurate. Such a policy evaluation considering only a single Q-value could still result in a significant bias. For a DRL agent with multiple critics in a designated task, although the critics have different initial parameters, they share the same optimization target and will converge to near-identical parameters as they approach optimal performance. However, during early training stages, the Q-values they produce can vary significantly, and identifying the most accurate one is challenging without a reference, such as a well-trained critic in the designated task. In this context, compared with favoring and relying solely on a single Q-value, our APE2 algorithm considers the average performance of multiple critics and uses the average Q-value of them to mitigate the bias introduced by any single critic similar to the approach in \cite{Wu2018}.}

Besides $Q_{\rm{LTR}}$, APE2 also incorporates the immediate return $R_{\rm{IR}}$ as part of the policy evaluation
\begin{align}
R_{\rm{IR}} = \sum\nolimits^{\rm H}_{h=1}r_{t-1+h}\label{eq4.4}
\end{align}
where $t$ is the current time step. $\rm H$ is the horizon, indicating the number of time steps over which the reward is considered. 

It is worth noting that the proposed parameterized environment allows for the efficient calculation of rewards without real interaction with a simulated or real-world platform. Since the obstacles and the robot are described using mathematical models, including kinematic and spatial geometric equations, both the resulting state ${\bm s}^\prime$ and the reward ${\bm r}$ can be computed analytically for any given state-action pair $({\bm s},{\bm a})$.

Considering both the long-term and immediate returns, the proposed hybrid policy evaluation is defined as
\begin{align}
V_{\rm{HPE}} = \big[\eta \;\big\lvert\; 1-\eta \big] \big[Q_{\rm{LTR}} \;\big\lvert\; R_{\rm{IR}} \big]^{\rm T} \label{eq4.5}
\end{align}
where $\eta= {\rm clip}(t_c/{\rm T},0,1)$. $t_c$ is the cumulative number of policy optimization, and $\rm T$ is a constant. $\eta$ is clipped to [0,1].

The proposed APE2 algorithm evaluates each action candidate in $A_t$ by the hybrid policy evaluation, and the optimal action is defined as
\begin{align}
\bar{\bm a}_t & = \underset{{\bm a} \in {A_t}}{\arg\max}\,{V_{\rm HPE}}\big\lvert_{{\bm s}_t, {\bm a}_t} \label{eq4.6}
\end{align}

\emph{Remark 5: }{Although $Q_{\rm{LTR}}$ provides a more unbiased, long-term return for policy evaluation as reported in \cite{FuXiaokuan,Wu2018}, the estimation is quite inaccurate in the early training stages due to insufficient training. In contrast, $R_{\rm{IR}}$ is accurate but shortsighted, focusing merely on the immediate return the agent receives in $\rm H$ time steps from the current time step. According to Eq. \eqref{eq4.5}, APE2 leverages the immediate return $R_{\rm{IR}}$ to provide a more accurate evaluation during the initial training phase. As training proceeds, the weight of $R_{\rm{IR}}$ decays to zero, transitioning the evaluation to rely solely on $Q_{\rm{LTR}}$. Hence, APE2 provides a more accurate and holistic evaluation throughout the training process. While a larger $\rm H$ can make $R_{\rm{IR}}$ closer to the long-term return, it also increases computational overhead. In practice, setting a small $\rm H$ and let it provide an accurate evaluation in the early training stage is sufficient, since the critics progressively learn to approximate the long-term return defined in Eq. \eqref{eq01} as training proceeds. Moreover, APE2 allows efficient generation of diverse action candidates by Eq. \eqref{eq4.2}, which enhances the exploration ability compared to the multi-critic algorithm in \cite{Wu2018}.}

\subsection{Network Training}
Given an interaction data $({\bm s},{\bm a},r,{\bm s}^\prime)$, the temporal difference error \cite{Fengkang2024} of each online critic in APE2 is defined as
\begin{align}
L_k = r + {\xi} {Q_{k}^\prime} \Big( {\bm s}^\prime,\mu^\prime \big({\bm s}^\prime;\theta^{\mu^\prime} \big)|\theta_{k}^{Q^\prime} \Big) - Q_{k}\big({\bm s}, {\bm a}|\theta_{k}^{Q} \big)    \label{eq4.7}
\end{align}
where $\theta_{k}^{Q^\prime}$ is the parameter of the target critic network $Q_{k}^\prime$. $\theta^{\mu^\prime}$ is the parameter of the target actor network $\mu^\prime$.

Based on Eq. \eqref{eq4.7}, the proposed APE2 optimizes each online critic $Q_k$ by the hybrid loss $L_{{\rm{H,}}k}$ defined as
\begin{align}
L_{{\rm{H,}}k} = {\omega_1} L_{k}^{2}+\Big(\frac{1-{\omega_1}}{\rm K}\sum\nolimits^{\rm K}_{k=1}L_{k}^{2}\Big)+{\omega_2}(\Delta Q)^2 
\label{eq4.8}
\end{align}
where $\omega_1$ and $\omega_2$ are the weight coefficients. $(\Delta Q)^2=({Q_k}-{Q_{\rm{LTR}}})^2$ is an auxiliary punishment that limits the difference among multiple critics \cite{FuXiaokuan}.

The proposed APE2 optimizes the online actor $\mu$ by
\begin{align}
\nabla_{\theta^\mu}J \approx \mathbb{E}_{\mu}\Big[\nabla_{{\bm a}}Q_{\rm{LTR}}  \nabla_{\theta^\mu} \mu\big({\bm s};\theta^\mu \big) \Big]   \label{eq4.10}
\end{align}

Additionally, the target actor $\mu^\prime$ and each target critic $Q_{k}^\prime$ are optimized by soft update \cite{FengkangASME}.

\section{Expert Data Diffusion and Utilization}
\label{sec:V}
In this section, we discuss how a diffusion model is leveraged to generate a large-scale trajectory dataset from very limited human demonstrations. Then, two methods for utilizing the diffused dataset are explored to train a robust motion planning policy.


\begin{figure}[!t]
\centering
\includegraphics[width=0.85\columnwidth]{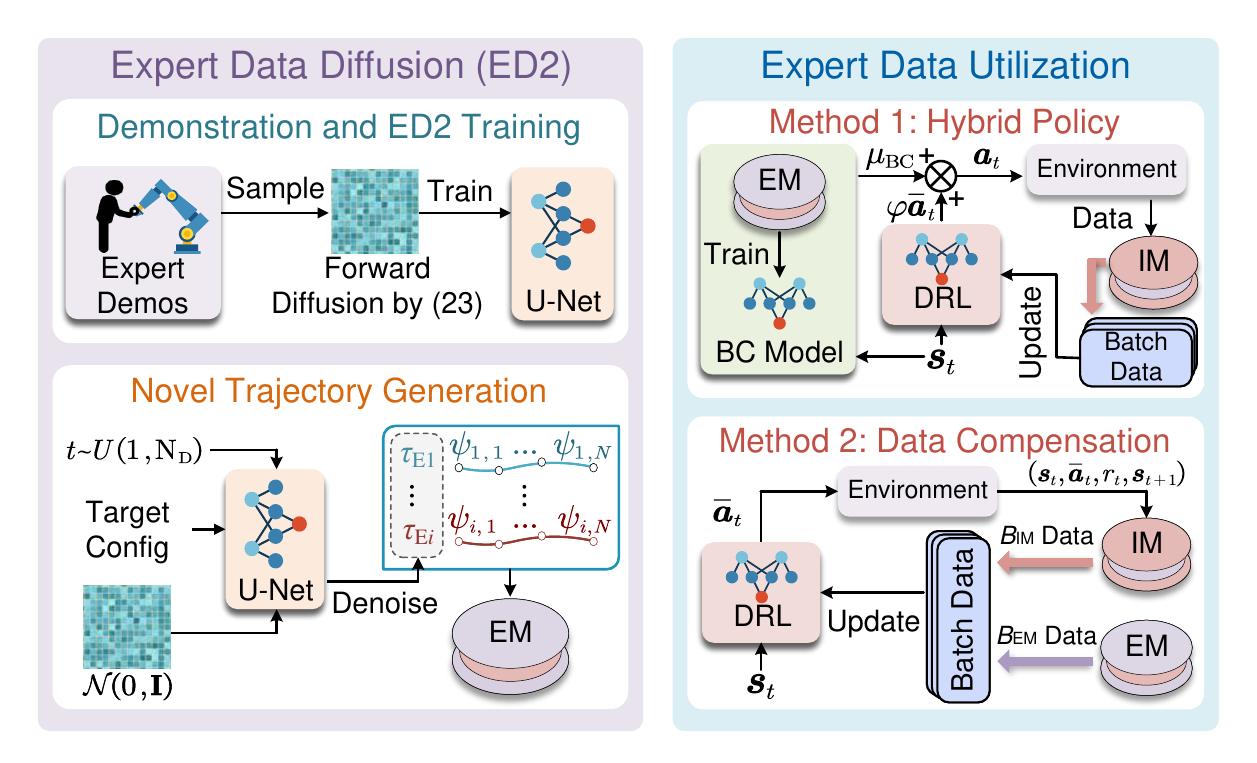}
\caption{Frameworks of the ED2 model and two expert data utilization methods. Given a target area, the ED2 model can be efficiently trained using a very limited set of expert demonstrations. The trained model can generate a large number of novel trajectories. The generated expert data can be utilized in two ways: the hybrid policy method creates a synergy between behavior cloning and DRL but may not yield optimal trajectories, while the data compensation method enhances DRL training and ensures learning optimal policies.}
\label{Fig5.1}
\end{figure}

\subsection{Expert Data Diffusion (ED2) from Limited Demonstrations}
\label{sec:VA}
As shown in Fig. \ref{Fig5.1}, the diffusion model $\epsilon_\theta$ of ED2 transforms a trajectory $\tau$ from the initial data distribution $\tau_0 \sim {\bm q}(\tau_0)$ into white Gaussian noise by running a Markovian forward diffusion process
\begin{equation}
{\bm q}(\tau_t|\tau_{t-1}, t) = \mathcal{N} (\tau_t; \sqrt{1 - \beta_t}\tau_{t-1}, \beta_t\mathbf{I})
\label{eq5.1}
\end{equation}
where $t \sim U(1, {\rm N_D})$ is the diffusion time step with uniform distribution $U$, $\beta_t$ is the noise scale at time step $t$ \cite{Ho2020ddpm}, and $\rm N_D$ is the number of diffusion steps. The value of $\rm N_D$ is determined empirically based on the trajectory length of the expert demonstrations. Assuming the data $\tau_0$ lives in Euclidean space, the distribution of the diffusion process at time step $t$ is Gaussian and can be written in closed-form as
\begin{equation}
{\bm q}(\tau_t|\tau_0, t) = \mathcal{N} (\tau_t; \sqrt{\bar{\alpha}_t}\tau_0, (1 - \bar{\alpha}_t)\mathbf{I})
\label{eq5.2}
\end{equation}
with $\alpha_t = 1 - \beta_t$ and $\bar{\alpha}_t = \prod_{i=1}^t \alpha_i$. This allows sampling $\tau_t$ without running the forward diffusion process \cite{Ho2020ddpm}.

The inverse (denoising) process of the proposed ED2 transforms Gaussian noise back to the data distribution through a series of denoising steps $p(\tau_{t-1}|\tau_t, t)$. ED2 approximates this posterior distribution with a parametrized Gaussian
\begin{equation}
p_{\theta} (\tau_{t-1}|\tau_t, t) = \mathcal{N} (\tau_{t-1}; \mu_t = \mu_{\theta} (\tau_t, t), \Sigma_t)
\label{eq5.3}
\end{equation}
For simplicity, only the mean of the inverse process is learned, and the covariance is set to $\Sigma_t = \sigma^2_t \mathbf{I} = \tilde{\beta}_t\mathbf{I}$, with $\tilde{\beta}_t = \beta_t(1 - \bar{\alpha}_{t-1})/(1 - \bar{\alpha}_t)$. Ho \textit{et al.} \cite{Ho2020ddpm} suggested that, rather than directly learning the posterior mean, the noise term $\epsilon$ could be learned instead via a simplified loss function
\begin{equation}
L(\theta) = \mathbb{E}_{t,\epsilon,\tau_0} [\|\epsilon - \epsilon_{\theta} (\tau_t, t)\|^2]
\label{eq5.4}
\end{equation}
with $t \sim U(1, \rm N_D)$, $\epsilon \sim \mathcal{N} (0, \mathbf{I})$, $\tau_t = \sqrt{\bar{\alpha}_t}\tau_0 + \sqrt{1 - \bar{\alpha}_t}\epsilon$, and  $\tau_0 \sim {\bm q}(\tau_0)$. We encode the diffusion model over trajectories with a temporal U-Net \cite{janner2022diffuser}, which has proven to be a reasonable architecture for diffusion models over trajectories. 


\emph{Remark 6: }{Algorithm \ref{alg:train_diffusion} presents the pseudocode for training the proposed ED2 model using a very limited set of collision-free trajectories obtained by an expert agent. Due to the small sample size, the training process is highly efficient. The trained model is capable of generating a large number of novel trajectories without the necessity of environmental exploration. Although these trajectories are non-optimal, they enable the manipulator to move to various poses within the designated workspace. Moreover, by leveraging the non-optimal dataset, DRL can explore and learn an optimal motion planning policy more efficiently. Compared to existing methods that purely rely on human demonstrations \cite{FengkangTII} or traditional motion planners \cite{XiangjianTII} to obtain substantial expert data, our method significantly reduces manual effort and offers better stability.}


\IncMargin{1em}
\begin{algorithm} \SetKwData{Left}{left}\SetKwData{This}{this}\SetKwData{Up}{up} \SetKwFunction{Union}{Union}\SetKwFunction{FindCompress}{FindCompress} \SetKwInOut{Input}{input}\SetKwInOut{Output}{output}	
     Collision-free trajectories $\mathcal{D}$, Diffusion model $\epsilon_\theta$, noise schedule terms $\bar{\alpha}_t$;\\
     \While{training is not finished}{
     Sample a batch of trajectories: $\tau_0 \sim \mathcal{D}, \ \epsilon \sim \mathcal{N}(0, I), \ t \sim \mathcal{U}(1, \rm N_D)$; \\
     Compute the diffusion loss function: \\
     $\tau_t = \sqrt{\bar{\alpha}_t} \tau_0 + \sqrt{1 - \bar{\alpha}_t} \epsilon$;\\
     $\mathcal{L}(\theta) = \| \epsilon - \epsilon_\theta(\tau_t, t) \|^2$;\\
     Gradient update.
     }
\caption{Training the ED2 Model.}
\label{alg:train_diffusion}
\end{algorithm}
\DecMargin{1em}



With the safety of the provided demonstrations assured, ED2 can be effectively trained to generate collision-free trajectories. The parameterized space verifies the trajectories generated by ED2 according to Eqs. \eqref{eq3.5} and \eqref{eq3.6}, ensuring that only collision-free trajectories are utilized. The trajectory $\tau_{\rm E}$ generated by the trained ED2 model $\epsilon_\theta\prime$ are converted to $\psi({\bm s}, {\bm a}, r, {\bm s}^{\prime})$ format and stored in the expert memory (EM). This process is very efficient using our parameterized task space. We then explore and compare two methods for utilizing the diffused dataset.

\subsection{Synergizing BC and DRL for Motion Planning}
\label{sec:VB}
In existing works, a popular approach to facilitate DRL training is the formation of a hybrid policy by combining DRL with behavior cloning \cite{LHF2023,Ma2021}. As shown in Fig. \ref{Fig5.1}, a BC model $\mu_{\rm{BC}}({\bm s}_t, \theta^{\mu}_{\rm {BC}})$ is developed to learn the mapping between state-action pairs at time $t$ (i.e., how $s_t$ maps to $a_t$) based on the diffused dataset generated by the proposed ED2 model. The loss function is defined as $L_{\rm{BC}}=\big({{\bm a}_{\rm E}}-\mu_{\rm{BC}}({\bm s}_{\rm E}; \theta^{\mu}_{\rm {BC}})\big)^2$, where $({\bm s}_{\rm E},{\bm a}_{\rm E})$ is the labeled action-state pair stored in EM. Since the diffusion step of ED2 is $\rm N_D$, the generated expert trajectory consists of $\rm N_D$ joint configuration sequences. Hence, the BC model trained from these trajectory data can only plan motions with fixed $\rm N_D$ steps.

For the first $\rm N_D$ steps, the action is generated by a hybrid policy wherein the BC model functions as the base policy, and the DRL model serves as the residual policy with a weight $\varphi$
\begin{equation}
{\bm a}_t=\pi\left({\bm s}_t\right)=\mu_{\rm{BC}}({\bm s}_t, \theta^{\mu}_{\rm {BC}})+\varphi\bar{\bm a}_t
\label{eq5.5}
\end{equation}

The weighted sum shown in Eq. \eqref{eq5.5} is a common approach to form a hybrid policy \cite{Ma2021}, which can create a synergy between BC and DRL. The weight $\varphi$ should be relatively small. In the hybrid policy, BC provides a base policy to ensure a reasonable performance from the outset, while DRL uses actions with small amplitudes to refine the actions suggested by the base policy, making it more robust to noise disturbances. With the inclusion of the base policy, the tool center point is positioned near the random target after the first $\rm N_D$ steps. Thus, we define ${\bm a}_t=\varphi\bar{\bm a}_t$ for subsequent steps, as using smaller actions allows the agent to effectively explore around the target area and learn to reach the randomly assigned target points. Additionally, DRL endows the agent with the ability to keep the tool center point within allowable errors after it reaches the target, even without a termination signal.

\subsection{Diffused Expert Data Compensation Mechanism}
\label{sec:VC}
Although the hybrid policy proposed in Section \ref{sec:VB} can improve both BC and DRL models, its motion planning performance still largely depends on the quality of the demonstrations. Given that neither the human demonstrations nor the diffused dataset are optimal, the hybrid policy often falls short in planning an optimal trajectory.

As shown in Fig. \ref{Fig5.1}, instead of directly cloning the behavior, another popular method is adopting an expert data compensation (DC) mechanism \cite{FengkangTII,Fengkang2024,Ball2023} in which the batch data for training DRL agent are sampled from both the interaction memory (IM) pool, which stores the interaction data, and the EM, which stores the diffused expert data. Define $B_{\rm IM}$ as the size of data sampled from IM
\begin{equation}
B_{\rm IM}= B - B_{\rm EM}
\label{eq5.6}
\end{equation}
where $B$ is the batch size. When the number of interaction data collected in IM is fewer than $B$, the value of $B$ is adjusted to match the total number of interaction data available. $B_{\rm EM}={\rm clip}(\lfloor t/{\rm T_{B}} \rfloor,0,B_{\rm EM}^{\rm max})$ is the size of data sampled from EM, and ${\rm T_{B}}$ is a constant. Unlike the approach in \cite{Ball2023}, where $B_{\rm EM}$ is set as a constant, we allow $B_{\rm EM}$ to increase progressively from zero to a maximum value as training proceeds. This is to balance $B_{\rm EM}$ with $B_{\rm IM}$, as there is limited interaction data collected at the beginning of training.

While the agent continuously uses interaction data to update the policy, the proposed DC mechanism gradually supplies additional high-quality data. It is worth noting that although we ensure that the trajectories generated by ED2 are collision-free, the quality of the data in EM does not affect DRL in learning optimal policies. This is because the core capability of DRL models lies in learning optimal planning through experience data in both EM and IM, with the latter constantly providing new interaction data. Previous studies \cite{Ball2023} have demonstrated that either small number of high-quality expert demonstrations, or even low-quality but high-coverage demonstrations, can accelerate DRL training. Consequently, compared to the hybrid policy, DC mechanism not only accelerates the policy searching and learning process of DRL but also ensures that the agent can learn optimal policies, since the action is no longer restricted by BC. 


\section{Overall Implementation of URPlanner}
In this section, the overall framework of our URPlanner is introduced, which integrates the proposed parameterized space, UOAR, APE2 algorithm, and ED2 model.

\begin{figure}[!t]
\centering
\includegraphics[width=3.2in]{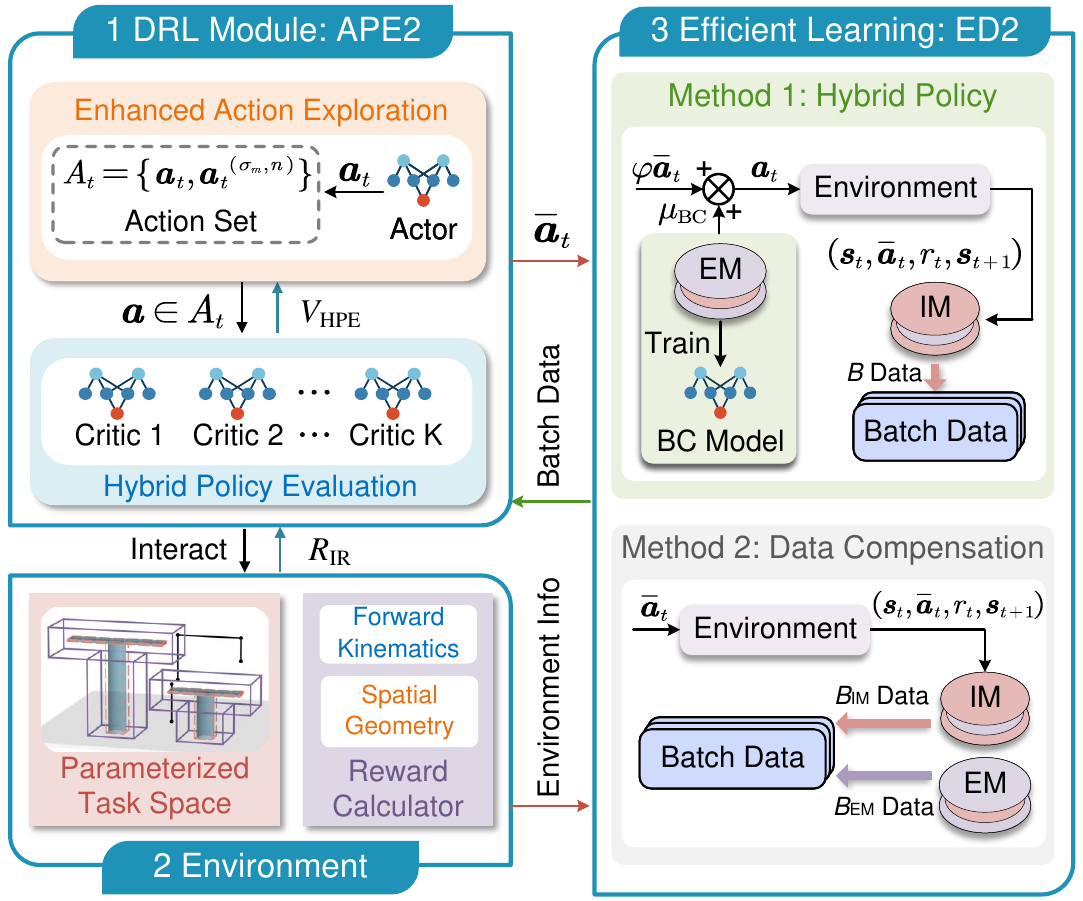}
\caption{Overall framework of URPlanner. The proposed URPlanner consists of three modules: a DRL module where APE2 is applied, an environment module for the parameterized space and reward calculation, and an efficient policy learning module for deploying the ED2 model.}
\label{Fig6.1}
\end{figure}

As shown in Fig. \ref{Fig6.1}, the proposed URPlanner consists of three modules: a DRL module where APE2 is applied, an environment module for the parameterized space and reward calculation, and an efficient policy learning module for deploying the ED2 model. Before training APE2, a certain number of expert trajectories are generated by ED2, converted to $({\bm s}, {\bm a}, r, {\bm s}^{\prime})$ format, and stored in the EM. If using the hybrid policy, a BC model is trained based on the expert data. Algorithm \ref{Algorithm2} provides an overall flow of the URPlanner.


\IncMargin{1em}
\begin{algorithm} \SetKwData{Left}{left}\SetKwData{This}{this}\SetKwData{Up}{up} \SetKwFunction{Union}{Union}\SetKwFunction{FindCompress}{FindCompress} \SetKwInOut{Input}{input}\SetKwInOut{Output}{output}
    Initialize APE2 with an actor and $\rm K$ critics;\\
    Initialize IM and EM, and fill up EM by ED2;\\
    \If{use hybrid policy}{
    Train a BC model $\mu_{\rm{BC}}$ based on EM;
    }   
    \For{each episode}{  
	Randomly reset the environment.\\
	\For{each time step $t$}{
        Generate a raw action  ${\bm a}_t$ by Eq. \eqref{eq4.1};\\
	Explore an action set $A_t$ by Eq. \eqref{eq4.2};\\
        Estimate the value $V_{\rm HPE}$ by Eq. \eqref{eq4.5};\\
        Select the optimal action $\bar{\bm a}_t$ defined by Eq. \eqref{eq4.6};\\
        \If{use hybrid policy}{
        Execute action by Eq. \eqref{eq5.5};
        }
        \textbf{else:} Execute $\bar{\bm a}_t$;\\
	Calculate the reward ${\bm r}_t$ by Eq. \eqref{eq3.10};\\
        Store $({\bm s}_t, {\bar {\bm a}}_t , {\bm r}_t, {\bm s}_{t+1})$ to IM;\\
        \If{use data compensation mechanism}{
        Sample $B$ data from IM and EM;
        }
        \textbf{else:} Sample $B$ data from IM only;\\
        Update online networks by Eqs. \eqref{eq4.8}-\eqref{eq4.10};\\
        Soft update target networks;\\
	\If{{the task is implemented}}{
       	Break.}  
 	 }   
}  
 	 	  \caption{Overall implementation of URPlanner.}
 	 	  \label{Algorithm2}
 	 \end{algorithm}
 \DecMargin{1em}

\section{Experimental Results}
In this section, the APE2 algorithm, UOAR, and ED2 strategy of the proposed URPlanner are experimentally verified against various state-of-the-art methods. We then compare URPlanner with existing learning-based and traditional planners. It is important to note that URPlanner is trained in the proposed parameterized space. In environments with static obstacle layouts, this parameterized space enables all necessary information for policy training, such as the resulting state and reward after executing an action, to be computed analytically without interacting with a simulated or real-world robot platform.

\subsection{Experimental Setups}
\begin{figure}[!t]
\centering
\vspace{0cm} 
\setlength{\belowcaptionskip}{-0.3cm} 
\includegraphics[width=3in]{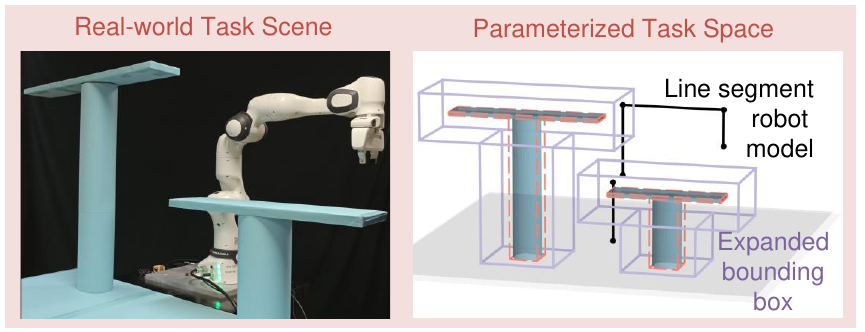}
\caption{Real-world scene and the corresponding parameterized space.}
\label{Fig7.1}
\end{figure}

The experiments in this section are conducted based on the task scene shown in Fig. \ref{Fig7.1}. The aim is to train a motion planning policy for a 7-DoF Franka Emika Panda robot, enabling it to reach randomly given goal poses above the table without any collision. The algorithms are trained on a laptop equipped with AMD Ryzen 9 5900HX CPU @ 3.30GHz and 32.0GB RAM, with the main parameters shown in Table \ref{TABLE I}. The network architecture for the actor and behavior cloning is \textit{state dimension}$\times$256(ReLu)$\times$256(ReLu)$\times$\textit{action dimension}(Tanh). The network architecture for the critics is \textit{state+action dimension}$\times$256(ReLu)$\times$256(ReLu)$\times$1. The dimension of state and action varies for manipulators with different DoFs. The structure of the U-Net in ED2 model can be found in \cite{janner2022diffuser}. The generated trajectory, represented as joint configuration sequences, is transmitted to the Franka robot via MoveIt, which subsequently sends actuation commands to the robot's joints. To control the robot's speed, the maximum velocity and acceleration scaling factors are set to 0.25.


\begin{table}[!t]
\centering
\vspace{0cm} 
\setlength{\abovecaptionskip}{0cm} 
\setlength{\belowcaptionskip}{0cm} 
\renewcommand{\arraystretch}{1.1}
\caption{Parameter Setups}
\label{TABLE I}
\setlength{\tabcolsep}{1.5mm}{
\begin{tabular}{cc|cc|cc}
\Xhline{0.7pt}
Parameter          & Value     & Parameter         & Value  & Parameter   & Value     
 \\ \hline
learning rate    & 1e-3    & $\zeta$     & 1     & $\omega_1$   & 0.6  \\
memory capacity  & 6e4     & $\rm M$     & 2     & $\omega_2$\  & 0.1  \\
soft update rate & 0.01    & $\rm N$     & 3     & $\rm N_D$    & 80   \\
$B$          & 64      & ${\rm T}$   & 2e5   & $\varphi$    & 0.1  \\
$\xi$        & 0.98    & H           & 1     &${\rm T}_{\rm B}$   & 2e3  \\
\Xhline{0.7pt}
\end{tabular}}
\end{table}

\subsection{Training Performance Evaluation}
\label{Section_VII.B}
\subsubsection{Evaluation of UOAR}
\label{Section_VII.B1}

\begin{table}[!t]
\centering
\vspace{0cm} 
\setlength{\abovecaptionskip}{0cm} 
\setlength{\belowcaptionskip}{0cm} 
\renewcommand{\arraystretch}{1.3}
\caption{Training Performances of DDPG Algorithm with Different Obstacle Avoidance Rewards}
\label{TABLE 7.2}
\setlength{\tabcolsep}{1.2mm}

\begin{threeparttable}
\begin{tabular}{cccccc}

\Xhline{0.7pt}
  &
  \multicolumn{4}{c}{\multirow{1.4}{*}{\textbf{Success Rate Per 250 Episodes} (\%)}} &
  \multicolumn{1}{c}{\multirow{1.4}{*}{\textbf{Execution Time} (s)}}
\\

\specialrule{0em}{1pt}{1pt} 
\cmidrule(r){2-5} \cmidrule(r){6-6}
\specialrule{0em}{1pt}{1pt} 
                &
  1-250         &  
  251-500       &
  501-750       &
  751-1000      &
  per 300 steps
\\
\specialrule{0em}{1pt}{1pt} 
\hline
\specialrule{0em}{1pt}{1pt} 
  CAR            &
  ${0.0}\:_{0.0}^{0.0}$          &
  ${0.0}\:_{0.0}^{0.0}$         &  
  ${32.3}\:_{0.0}^{77.2}$         &
  ${95.8}\:_{88.8}^{99.6}$     &  
  ${72.40}\:_{71.76}^{73.86}$
\\
  COR            &
  ${0.0}\:_{0.0}^{0.0}$          &
  ${6.1}\:_{0.0}^{18.0}$         &  
  ${46.6}\:_{0.8}^{66.4}$         &
  ${86.4}\:_{79.2}^{96.0}$     &  
  ${72.11}\:_{71.60}^{72.64}$
\\
  PLR            &
  ${0.0}\:_{0.0}^{0.0}$          &
  ${0.0}\:_{0.0}^{0.0}$          &  
  ${49.1}\:_{26.0}^{76.0}$          &
  ${99.0}\:_{98.4}^{99.6}$     &  
  ${9.351}\:_{9.119}^{9.526}$         
\\
  UOAR           &
  ${0.0}\:_{0.0}^{0.0}$          &
  ${0.0}\:_{0.0}^{0.0}$         &  
  ${56.3}\:_{33.2}^{78.0}$       &
  ${96.4}\:_{94.4}^{98.8}$     &  
  {\textbf{0.776}}\:$_{0.768}^{0.785}$   
\\
\specialrule{0em}{1pt}{1pt} 
\Xhline{0.7pt}
\end{tabular}
\begin{tablenotes}[para,flushleft]
	\footnotesize
	\textbf{Annotation:} The data are in the format of $\rm{mean_{min}^{max}}$. The success rate over a given episode interval $[a,b]$ is defined as $c/(b-a+1)\times100\%$, where $c$ represents the number of episodes in which the agent successfully completes the task. This metric reflects how fast the agent finds and learns a stable policy for completing the given task. The execution time counts in all the necessary procedures for a complete interaction between the agent and the environment, including the time for state observation, action calculation and execution, reward calculation, as well as communication. 
\end{tablenotes}
\end{threeparttable}
\end{table}

\begin{table}[!t]
\centering
\vspace{0cm} 
\setlength{\abovecaptionskip}{0cm} 
\setlength{\belowcaptionskip}{0cm} 
\renewcommand{\arraystretch}{1.3}
\caption{Quantitative and Qualitative Analyses of Different Obstacle Avoidance Rewards}
\label{TABLE 7.3}
\setlength{\tabcolsep}{1.2mm}

\begin{threeparttable}
    
\begin{tabular}{ccccc}

\Xhline{0.7pt}
  &
  \textbf{CAR}        &
  \textbf{COR}        &
  \textbf{PLR}        &
  \textbf{UOAR}

\\
\specialrule{0em}{1pt}{1pt} 
\hline
\specialrule{0em}{1pt}{1pt} 
  \textbf{\makecell[c]{Testing\\success rate}} &
  100.0\%          &
  100.0\%          &  
  100.0\%          &
  100.0\%
\\
\specialrule{0em}{1pt}{1pt} 
 \textbf{Trajectory length}        &
  $1.174_{1.136}^{1.203}$          &
  $1.103_{1.086}^{1.117}$          &  
  $1.186_{1.169}^{1.214}$          &
  $1.053_{1.036}^{1.072}$
\\
\specialrule{0em}{1pt}{1pt} 
  \textbf{\makecell[c]{Minimum distance\\independent}}            &
  $\times$          &
  $\times$          &  
  $\times$          &
  \checkmark
\\
  \textbf{Platform-agnostic}            &
  $\times$          &
  $\times$          &  
  \checkmark        &
  \checkmark
\\
 \textbf{Efficiency}             &
  very low          &
  very low          &  
       high         &
  very high         
\\
 \textbf{Practical value}  &
  normal                &
  normal                &  
  high                &
  very high  
\\
\specialrule{0em}{1pt}{1pt} 
\Xhline{0.7pt}
\end{tabular}
\begin{tablenotes}[para,flushleft]
	\footnotesize
	\textbf{Annotation:} The testing success rate is calculated based on 50 trials. The trajectory length is in the format of $\rm{mean_{min}^{max}}$. The practical value evaluates the time cost and effectiveness in real-world applications. 
\end{tablenotes}
\end{threeparttable}
\end{table}

\begin{figure*}[!t]
\centering
\includegraphics[width=6.2in]{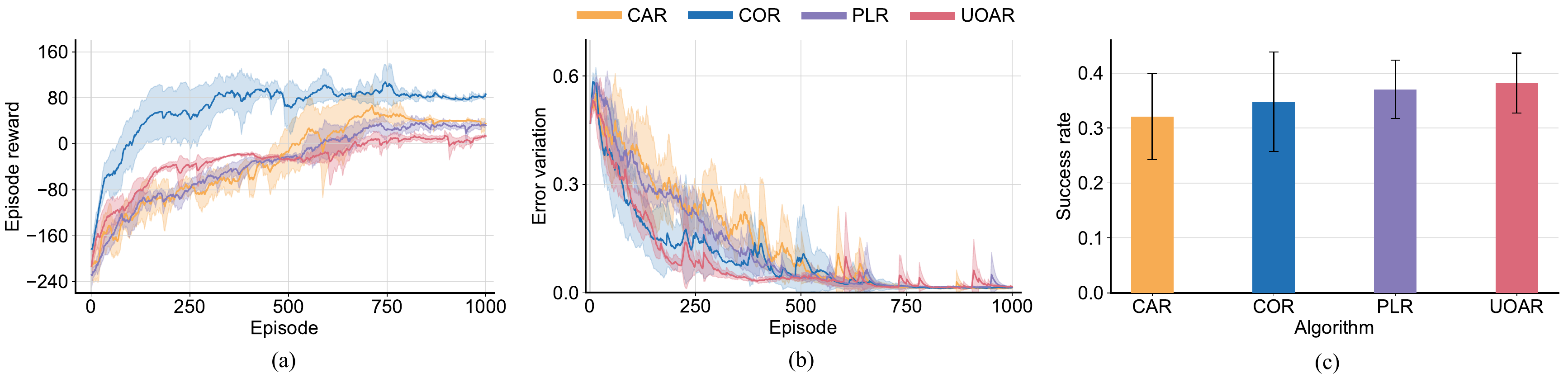}
\caption{Training performances of DDPG algorithms based on different obstacle avoidance rewards. The shaded region and the error bar represent the standard deviation of the average evaluation over four random seeds. (a) Episode reward. (b) Error variation. (c) Total training success rate.}
\label{Fig7.x1}
\end{figure*}

Firstly, we compare the training performance of the proposed UOAR with other obstacle avoidance rewards reported in recent research. For each method, we substitute only Eq. \eqref{eq3.8} of UOAR by other obstacle avoidance rewards. Referring to Eqs. (16)-(18) in \cite{Fengkang2024}, COR is a quadratic function based on the precise minimum distance (MD) between the manipulator and obstacles, consisting of an attraction component and a repulsive component. CAR is a linear function proportional to the MD, which was proposed in \cite{Sui2021} and modified in \cite{Fengkang2024}. Its setup can be found in Eq. (21) of \cite{Fengkang2024}. Both methods have been proven effective in collision-free robotic motion planning. In \cite{Fengkang2024}, COR and CAR are non-platform-agnostic methods, and the DRL agent is trained on CoppeliaSim.

It is worth noting that MD can also be approximately obtained in the proposed parameterized space through proper abstraction. This allows MD-based approaches, such as CAR and COR, to be designed in a platform-agnostic manner. Prior studies like \cite{CHEN202264} and \cite{Yang2022} simplified the acquisition of MD by calculating the distance between an obstacle's centre point and the manipulator's link. Based on this, they introduced a centre-point-to-line reward to learn obstacle avoidance. However, this method oversimplifies obstacles by modeling them as mass points, leading to a significant loss of spatial information. As a result, it struggles to handle the complex task scenarios addressed in our work. Thus, we make some improvements to it and design a points-to-line reward (PLR). PLR refines the bounding box representation by subdividing its surface into multiple points at equal intervals. The MD is then calculated by identifying the shortest distance between all points on the bounding box and the manipulator's line segments. This MD is used by PLR with the same reward function as CAR. Both PLR and the proposed UOAR are platform-agnostic and trained in the parameterized space.


In this experiment, DDPG is trained using different obstacle avoidance rewards. As shown in Table \ref{TABLE 7.2}, we use the success rate in training to reflect how fast the DRL agent with different obstacle avoidance rewards finds and learns a stable policy for completing the given task. Between the 1st to 750th training episodes, the success rates (averaged over four random seeds) of the four methods are relatively low, suggesting that the agent has not yet learned a stable policy for completing the tasks and requires further training. In contrast, between the 751st and 1000th episodes, the training success rates of CAR, PLR, and UOAR reach nearly 100\%, indicating that the policies have converged to a desired state, and the training process can be terminated.

Although COR achieves a success rate of 6.1\% between the 251st to 500th episodes, compared to zero for the other three reward functions, it performs slightly worse in the subsequent training period. According to the success rates in Table \ref{TABLE 7.2} and Fig. \ref{Fig7.x1}(c), PLR and the proposed UOAR exhibit similar performance in learning a stable policy for completing the tasks, and both of them slightly outperform CAR and COR.

Besides the efficiency in converging to a stable policy, we also compare the execution time required for training across different methods to evaluate their time cost. In this work, we propose a DRL-based motion planning paradigm that is highly cost-effective in both training and deployment. This efficiency arises from the proposed parameterized space and the MD-independent reward function, UOAR. 

First, our platform-agnostic method uses parameterized space to represent the environment, enabling all necessary information for training the DRL agent to be computed analytically. In contrast, existing DRL-based motion planners, such as those in \cite{XiangjianTII, FengkangASME,FengkangTII,Fengkang2024,R11,CHEN202264,Ma2021}, rely on interactions with specific simulated or real-world robotic platforms to obtain training data. 
These methods incur significant time costs due to communication between the agent and the platform to send states and actions, and the waiting time for action execution to retrieve the resulting state and reward. As shown in Table \ref{TABLE 7.3}, CAR and COR are non-platform-agnostic methods, while PLR and UOAR are platform-agnostic methods based on the proposed parameterized space. Table \ref{TABLE 7.2} indicates that CAR and COR take over 70 s to complete 300 interactions, whereas PLR and UOAR require only 9.351 s and 0.776 s, respectively. This substantial reduction in execution time demonstrates the significant advantage of our platform-agnostic approach in minimizing training time.

Second, the proposed UOAR strikes a balance between simplifying reward calculations and preserving sufficient spatial information about the manipulator and obstacles, while also maintaining satisfactory training success rates. Existing DRL-based planners rely heavily on computing the MD between the robot and obstacles to calculate rewards, which is more time-intensive. The PLR and UOAR in Table \ref{TABLE 7.2} provide a fair comparison on the time cost for calculating rewards, as both methods are trained in the parameterized space using the same DRL algorithm, differing only in their reward calculation methods. In the scenarios with four obstacles, PLR requires 9.351 s for 300 interactions, while UOAR requires only 0.776 s. This difference, stemming solely from the reward computation process, underscores the advantage of UOAR in significantly reducing the time cost of training. Moreover, our method shows satisfactory scalability in terms of computational efficiency. In the scenario with eight obstacles, the execution time (for 300 interactions) increases to ${0.799}\:_{0.786}^{0.824}$ s for UOAR and ${16.87}\:_{16.54}^{17.04}$ s for PLR. In an environment with sixteen obstacles, UOAR completes the same number of interactions in ${0.861}\:_{0.840}^{0.886}$ s, while PLR requires ${32.77}\:_{32.45}^{33.36}$ s. These results clearly show that as the number of obstacles increases, the computational advantage of UOAR becomes even more pronounced.
 
Additionally, more analysis of the four obstacle avoidance rewards is provided in Table \ref{TABLE 7.3}. In conclusion, the proposed UOAR demonstrates superior practical value than other counterparts, as it is MD-independent, platform-agnostic, and highly efficient in both training and deployment. Although UOAR relies on certain abstractions of real-world environment, such as padding obstacle's bounding box, it can be seen that a proper expansion does not result in an increase in trajectory length. In Section \ref{sec:VII-C}, more comparisons regarding the quality of trajectories planned by these rewards will be presented in real-world applications. In the subsequent experiments, all the algorithms will be trained using UOAR unless otherwise specified.

\subsubsection{Evaluation of APE2}

\begin{figure*}[!t]
\centering
\includegraphics[width=6.2in]{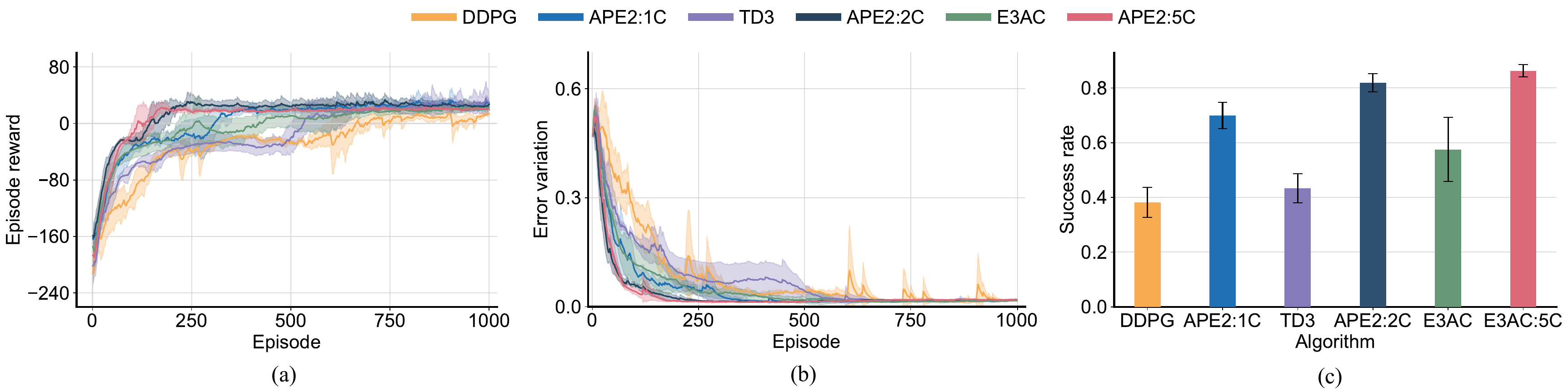}
\caption{Training performances of different DRL algorithms with UOAR. The shaded region and the error bar represent the standard deviation of the average evaluation over four random seeds. (a) Episode reward. (b) Error variation. (c) Total training success rate.}
\label{Fig7.2}
\end{figure*}

Secondly, the proposed APE2 algorithm is applied to various DRL benchmarks, including DDPG, TD3, and E3AC, to evaluate its superiority and universality.  

DDPG is a typical DRL algorithm based on deterministic policy gradient, which performs well in handling high-dimensional and continuous spaces. However, it often exhibits unsatisfactory learning efficiency and stability, especially in complex tasks such as collision-free motion planning for redundant manipulators. As shown in Table \ref{TABLE 7.4}, the training success rate is zero in the first 500 episodes. 
Although DDPG successfully learns a policy to complete the tasks during the 501st to 1000th episodes, Figs. \ref{Fig7.2} (a)-(b) reveal that the policy does not converge until approximately the 650th episode. Specifically, episode rewards are a key metric for evaluating the training performance of DRL, reflecting how effectively the agent maximizes the reward signal over time. Notably, the episode rewards for DDPG remain relatively lower than those of its counterparts after convergence. Furthermore, significant reward fluctuations are observed around the 750th and 900th episodes. While minor fluctuations are normal post-convergence, substantial fluctuations often indicate instability in training. To quantify this, we calculate the standard deviation of episode rewards during training as a measure of reward fluctuation (RF). The RF for DDPG (averaged over four seeds) is 45.2 and is the largest among all compared methods.

\begin{table}
\centering
\vspace{0cm} 
\setlength{\abovecaptionskip}{0cm} 
\setlength{\belowcaptionskip}{0cm} 
\renewcommand{\arraystretch}{1.3}
\caption{Training Success Rate and Stability of Different DRL Algorithms}
\label{TABLE 7.4}
\setlength{\tabcolsep}{1.2mm}

\begin{threeparttable}
\begin{tabular}{cccccc}

\Xhline{0.7pt}
  &
  \multicolumn{4}{c}{\multirow{1.4}{*}{\textbf{Success Rate Per 250 Episodes (\%)}}} &
  \multicolumn{1}{c}{\multirow{3}{*}{\textbf{\makecell[c]{Reward\\Fluctuation\\(RF)}}}}
\\

\specialrule{0em}{1pt}{1pt} 
\cmidrule(r){2-5} 
\specialrule{0em}{1pt}{1pt} 
          &
  1-250   &  
  251-500 &
  501-750 &
  751-1000&
  
\\
\specialrule{0em}{1pt}{1pt} 
\hline
\specialrule{0em}{1pt}{1pt} 
  DDPG           &
  $0.0\:_{0.0}^{0.0}$          &
  $0.0\:_{0.0}^{0.0}$          &  
  $56.3\:_{33.2}^{78.0}$         &
  $96.4\:_{94.4}^{98.8}$         &  
  $45.2\:_{34.3}^{52.4}$
\\
  APE2:1C        &
  $2.5\:_{0.0}^{10.0}$          &
  $79.0\:_{56.0}^{99.2}$         &  
  $99.4\:_{98.8}^{100.0}$         &
  $99.0\:_{98.0}^{100.0}$         &  
  $39.4\:_{36.3}^{44.8}$        
\\
\specialrule{0em}{1pt}{1pt} 
\hline
\specialrule{0em}{1pt}{1pt} 
  TD3            &
  $0.0\:_{0.0}^{0.0}$          &
  $0.2\:_{0.0}^{0.4}$          &  
  $74.2\:_{37.6}^{92.4}$         &
  $98.8\:_{96.8}^{100.0}$         &  
  $44.4\:_{39.7}^{49.7}$         
\\
  APE2:2C &
  $28.4\:_{7.2}^{44.0}$          &
  $100.0\:_{100.0}^{100.0}$        &  
  $99.6\:_{99.2}^{100.0}$         &
  $99.7\:_{99.2}^{100.0}$        &  
  $31.9\:_{27.7}^{34.7}$   
\\
\specialrule{0em}{1pt}{1pt} 
\hline
\specialrule{0em}{1pt}{1pt} 
  E3AC &
  $0.0\:_{0.0}^{0.0}$          &
  $40.4\:_{0.0}^{94.0}$         &  
  $90.5\:_{62.0}^{100.0}$        &
  $99.2\:_{98.8}^{99.6}$        &  
  $34.1\:_{27.5}^{40.1}$     
\\
  APE2:5C &
  \textbf{45.7}$\:_{36.8}^{60.4}$         &
  $100.0\:_{100.0}^{100.0}$        &  
  $99.9\:_{99.6}^{100.0}$        &
  $99.5\:_{99.2}^{100.0}$         &  
  \textbf{30.4}$\:_{26.7}^{34.8}$   
\\
\specialrule{0em}{1pt}{1pt} 
\Xhline{0.7pt}
\end{tabular}
\begin{tablenotes}[para,flushleft]
	\footnotesize
	\textbf{Annotation:} The data are in the format of $\rm{mean_{min}^{max}}$. APE:1C, APE2:2C, and APE2:5C differ only in the number of critics (1, 2, and 5, respectively). For a DRL algorithm, the RF represents the standard deviation of episode rewards during training, capturing the extent of their fluctuations. A smaller RF value indicates that the DRL algorithm is more stable in training. 
\end{tablenotes}
\end{threeparttable}
\end{table}

\begin{table}
\centering
\vspace{0cm} 
\setlength{\abovecaptionskip}{0cm} 
\setlength{\belowcaptionskip}{0cm} 
\renewcommand{\arraystretch}{1.3}
\caption{Testing Success Rate of Different DRL Algorithms}
\label{TABLE 7.4.2}
\setlength{\tabcolsep}{1.3mm}

\begin{threeparttable}
\begin{tabular}{ccccccc}

\Xhline{0.7pt}
     &
DDPG &
APE2:1C &
TD3  &
APE2:2C &
E3AC &
APE2:5C
  
\\
\specialrule{0em}{1pt}{1pt} 
\hline
\specialrule{0em}{1pt}{1pt} 
  seen           &
  100\%          &
  100\%          &  
  100\%         &
  100\%         &  
  100\%         &
  100\%
\\
  \makecell[c]{unseen\\($\leq\!5$ cm)}        &
  100\%          &
  100\%          &  
  100\%         &
  100\%         &  
  100\%         &
  100\% 
\\
  \makecell[c]{unseen\\(5 to 10 cm)}        &
  60\%          &
  100\%          &  
  58\%         &
  100\%         &  
  96\%         &
  100\%        
\\
\specialrule{0em}{1pt}{1pt} 
\Xhline{0.7pt}
\end{tabular}
\begin{tablenotes}[para,flushleft]
	\footnotesize
	\textbf{Annotation:} The testing success rate is calculated based on 50 trials. Two regions along the $y$-axis are tested for unseen goal positions: the region within 5 cm of the trained area, and the region 5 to 10 cm away from the trained area. In each region, 50 new target poses are uniformly distributed. 
\end{tablenotes}
\end{threeparttable}
\end{table}

By applying APE2 to DDPG, we create its APE2 variant, which has only one critic (denoted as APE2:1C). Hence, APE2:1C relies on a single Q-value to estimate the long-term return. APE2:1C explores diverse action candidates by Eq. \eqref{eq4.2}, evaluates their values by the proposed hybrid policy evaluation, and executes the optimal one according to Eq. \eqref{eq4.6}. Compared to DDPG, APE2:1C achieves a success rate (averaged over four random seeds) of 79.0\% during the 251st-500th training episodes and converges around the 350th episode. Moreover, the average RF during training is reduced to 39.4, indicating enhanced stability. As shown in Fig. \ref{Fig7.2}(a), the episode rewards are relatively stable after the policy converges, further demonstrating the improved training performance of APE2:1C compared to DDPG.

TD3 has two critics but uses only the minimum Q-value for policy optimization, which can mitigate the overestimation problem of DDPG. However, as shown in Table \ref{TABLE 7.4}, although the success rate (averaged over four seeds) during 501st-700th training episodes improves by 17.9\% compared to DDPG, the total success rate throughout training remains significantly lower than that of the three APE2 variants (APE2:1C, APE2:2C, and APE2:5C). Furthermore, TD3 assumes that the lower Q-value is always more accurate. Such a policy evaluation considering only a single Q-value could still result in a significant bias. Consequently, the RF of TD3 is very close to that of DDPG. In contrast, the APE2 variant of TD3, which employs two critics (denoted as APE2:2C), uses both Q-values during policy evaluation, resulting in notable training performance. As shown in Fig. \ref{Fig7.2}(a) and Table \ref{TABLE 7.4}, the policy converges in approximately 250 episodes and maintains a success rate of nearly 100\% thereafter. Additionally, the RF decreases to only 31.9, further demonstrating the stability and effectiveness of the proposed policy exploration and evaluation approach.

E3AC is one of the latest deterministic policy gradient-based algorithms, utilizing five Q-values for more unbiased policy evaluation. As shown in Table \ref{TABLE 7.4} and Fig. \ref{Fig7.2}(a), E3AC outperforms DDPG and TD3 in terms of success rate, convergence speed, and stability, both of which rely on a single Q-value. However, the total success rate throughout training remains significantly lower than that of the three APE2 variants. This limitation arises because, in the early stages of training, the Q-value estimation by the critics is still inaccurate due to insufficient training data and iterations. In contrast, the APE2 variant of E3AC, which incorporates five critics (denoted as APE2:5C), achieves a 45.7\% success rate during the 1st-250th episodes. Moreover, it exhibits the lowest RF at only 30.4. This outstanding performance can be attributed to the inclusion of the immediate return $R_{\rm IR}$ in policy evaluation, which enhances accuracy during the early training stages. 

Additionally, we evaluate the policies learned by the above DRL algorithms by testing them with seen and unseen goal poses after training is completed. As shown in Table \ref{TABLE 7.4.2}, all algorithms achieve a 100\% success rate for reaching 50 random goal positions within the trained target area without collision. To assess generalization capabilities, two regions along the $y$-axis are tested for unseen goal positions: the region within 5 cm of the trained area, and the region 5 to 10 cm away from the trained area. In each region, 50 unseen target poses are uniformly distributed for evaluation. Results show that for unseen goals within 5 cm of the trained area, all algorithms maintain a 100\% success rate. However, for unseen goals 5 to 10 cm away, benchmark algorithms such as DDPG and TD3 exhibit a significant drop in success rates. In contrast, the proposed APE2 algorithm, regardless of the number of critics, consistently achieves a 100\% success rate, demonstrating its superior generalization ability.

Based on the above discussion, it can be concluded that the proposed APE2 algorithm can effectively improve the performance of various DRL benchmarks by enhancing policy exploration and evaluation. According to Fig. \ref{Fig7.2}, APE2:1C, APE2:2C, and APE2:5C are the top three algorithms in terms of training success rate. Their performances are relatively more stable across different seeds as indicated by the narrower shaded region representing the standard deviation. By further comparing them, we observe that the training performance improves with an increased number of critics. Therefore, we use APE2:5C for subsequent experiments and simply denote it as APE2.

\subsubsection{Evaluation of ED2 and its utilization}
\label{sec:VIIB3}

\begin{table}
\centering
\vspace{0cm} 
\setlength{\abovecaptionskip}{0cm} 
\setlength{\belowcaptionskip}{0cm} 
\renewcommand{\arraystretch}{1.3}
\caption{Training Performances of Different Data Acquisition and Utilization Methods}
\label{TABLE 7.5}
\setlength{\tabcolsep}{1.5mm}

\begin{threeparttable}
\begin{tabular}{ccccc}
\Xhline{0.7pt}
  &
  \multicolumn{4}{c}{\multirow{1.4}{*}{\textbf{Success Rate Per 250 Episodes (\%)}}}
\\

\specialrule{0em}{1pt}{1pt} 
\cmidrule(r){2-5}
\specialrule{0em}{1pt}{1pt} 
  &
  1-250   &  
  251-500 &
  501-750 &
  751-1000
\\
\specialrule{0em}{1pt}{1pt} 
\hline
\specialrule{0em}{1pt}{1pt} 
  E3AC           &
  $0.0\:_{0.0}^{0.0}$          &
  $40.4\:_{0.0}^{94.0}$         &  
  $90.5\:_{62.0}^{100.0}$        &
  $99.2\:_{98.8}^{99.6}$           
\\
  E3AC:BC        &
  $16.8\:_{5.6}^{35.6}$         &
  $97.6\:_{96.0}^{100.0}$        &  
  $98.3\:_{97.2}^{99.2}$         &
  $97.0\:_{96.0}^{97.6}$         
\\
  E3AC:DC-RL    &
  $48.6\:_{45.2}^{50.8}$         &
  $99.9\:_{99.6}^{100.0}$         &  
  $99.5\:_{99.2}^{99.6}$         &
  $99.6\:_{98.8}^{100.0}$         
\\
  E3AC:DC-ED2    &
  $40.2\:_{36.8}^{42.8}$         &
  $99.9\:_{99.6}^{100.0}$         &  
  $99.3\:_{98.4}^{100.0}$         &
  $99.8\:_{99.2}^{100.0}$        
\\
\specialrule{0em}{1pt}{1pt} 
\hline
\specialrule{0em}{1pt}{1pt} 
  APE2           &
  $45.7\:_{36.8}^{60.4}$         &
  $100.0\:_{100.0}^{100.0}$        &  
  $99.9\:_{99.6}^{100.0}$        &
  $99.5\:_{99.2}^{100.0}$         
\\
  APE2:BC        &
  $53.0\:_{40.0}^{66.8}$         &
  $99.9\:_{99.6}^{100.0}$        &  
  $98.0\:_{97.6}^{98.4}$         &
  $99.1\:_{98.0}^{100.0}$         
\\
  APE2:DC-RL    &
  $67.8\:_{65.2}^{71.2}$         &
  $99.9\:_{99.6}^{100.0}$        &  
  $99.8\:_{99.6}^{100.0}$         &
  $99.3\:_{98.0}^{100.0}$         
\\
  APE2:DC-ED2    &
  $63.7\:_{61.2}^{66.8}$         &
  $99.9\:_{99.6}^{100.0}$        &  
  $99.8\:_{99.6}^{100.0}$         &
  $99.4\:_{99.2}^{99.6}$         
\\
\specialrule{0em}{1pt}{1pt} 
\Xhline{0.7pt}
\end{tabular}
\begin{tablenotes}[para,flushleft]
	\footnotesize
	\textbf{Annotation:} The data are in the format of $\rm{mean_{min}^{max}}$. All the algorithms achieve a 100\% testing success rate based on 50 trials. 
\end{tablenotes}
\end{threeparttable}
\end{table}

\begin{figure*}[!t]
\centering
\includegraphics[width=6.2in]{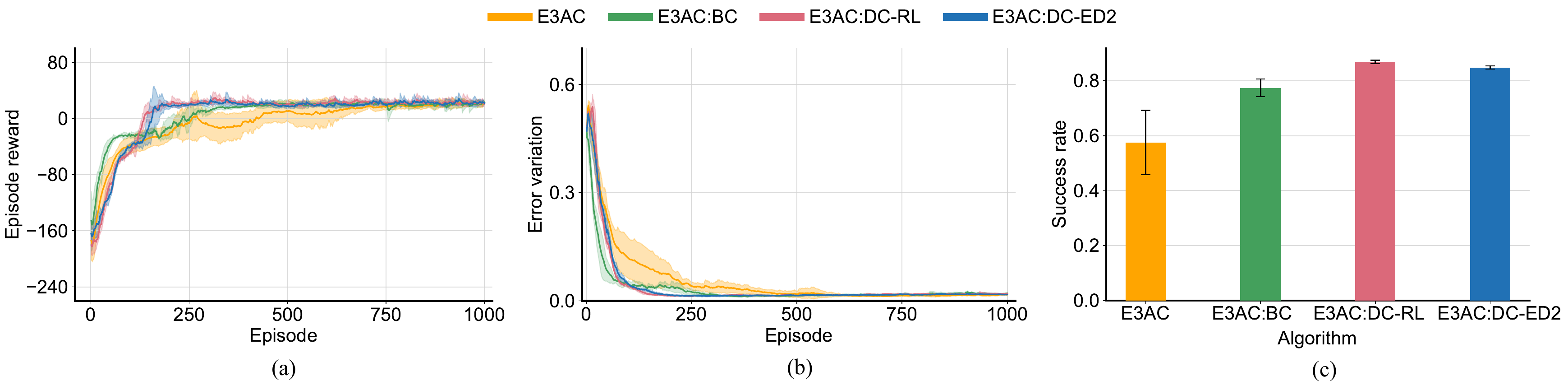}
\caption{Comparison of the hybrid policy and data compensation mechanism in facilitating the training of E3AC algorithm. The shaded region and the error bar represent the standard deviation of the average evaluation over four random seeds. (a) Episode reward. (b) Error variation. (c) Total training success rate.}
\label{Fig7.3}
\end{figure*}

\begin{figure*}[!t]
\centering
\includegraphics[width=6.2in]{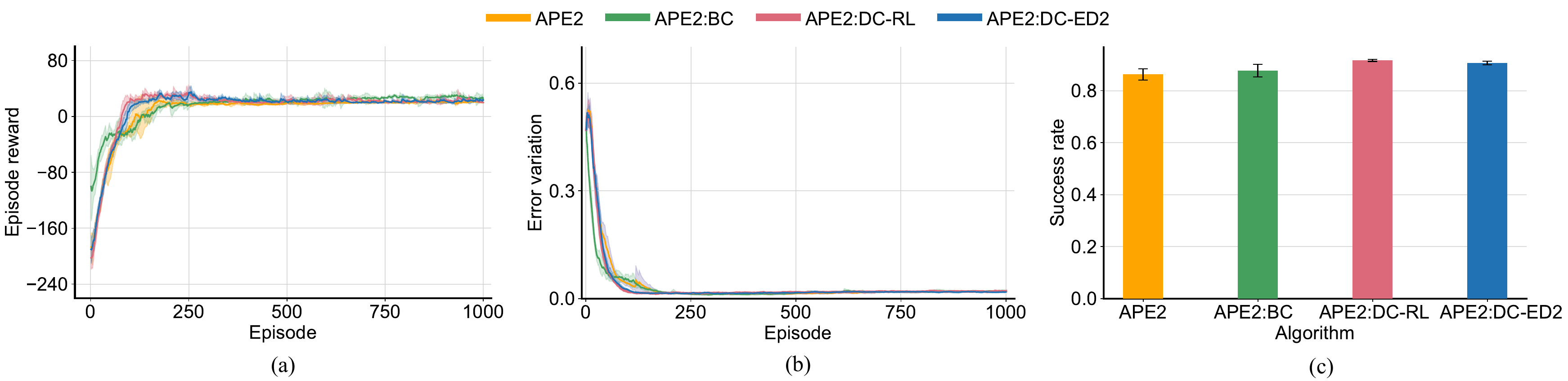}
\caption{Comparison of the hybrid policy and data compensation mechanism in facilitating the training of APE2 algorithm. The shaded region and the error bar represent the standard deviation of the average evaluation over four random seeds. (a) Episode reward. (b) Error variation. (c) Total training success rate.}
\label{Fig7.x2}
\end{figure*}

Lastly, we conduct a series of experiments to evaluate the impact of the diffused dataset (containing 800 expert trajectories) generated by the ED2 model on the training efficiency of DRL benchmarks. As shown in Table \ref{TABLE 7.5}, we perform a set of four experiments based on both E3AC and APE2. In each set, E3AC:BC and APE2:BC represent the hybrid policy discussed in Section \ref{sec:VB}, which creates a synergy between behavior cloning and DRL. DC-ED2 and DC-RL represent the expert data compensation mechanism discussed in Section \ref{sec:VC}, where the expert trajectory data in the expert memory are generated by two sources: the proposed ED2 model (DC-ED2) and a well-trained RL model (DC-RL). Thus, DC-RL contains optimal trajectory data, whereas DC-ED2 includes non-optimal data.

From the perspective of efficiency to train a stable policy, as shown in Table \ref{TABLE 7.5}, both E3AC and APE2 have already achieved satisfactory training success rates. By further combining a behavior cloning model trained using the diffused dataset generated by ED2 to form a hybrid policy, E3AC:BC and APE2:BC have promoted the success rate to 16.8\% and 53.0\% in the first 250 training episodes, respectively. Compared to the hybrid policy, the data compensation mechanism shows better improvements in training performance, achieving a success rate of 40.2\% for E3AC:DC-ED2 and 63.7\% for APE2:DC-ED2 during this training period. Specifically, it takes only about 100 training episodes for APE2:DC-ED2 to converge to a stable motion planning policy according to Fig. \ref{Fig7.x2}(a).

To further evaluate the effectiveness of the proposed ED2 in accelerating training with high-quality but non-optimal expert data, we use a well-trained RL model to provide optimal expert data for data compensation. As shown in Table \ref{TABLE 7.5}, both E3AC:DC-RL and APE2:DC-RL achieve the highest success rates in the first 250 training episodes, compared with the corresponding benchmarks. However, using DC-RL creates an idealized condition where the demonstrations are optimal, which is unrealistic given that the trajectories demonstrated by an expert are typically non-optimal. In contrast, ED2 is more realistic and practical, allowing DRL to search for and learn an optimal policy more efficiently with the aid of some high-quality but non-optimal expert data. As shown in Fig. \ref{Fig7.3}(a), and Fig. \ref{Fig7.x2}(a), the episode rewards of E3AC:DC-ED2 and APE2:DC-ED2 converge to values similar to E3AC:DC-RL and APE2:DC-RL, respectively. This indicates that DRL algorithms trained with the help of either optimal or non-optimal expert data can learn comparable policies. Furthermore, as depicted in Fig. \ref{Fig7.3}(c) and Fig. \ref{Fig7.x2}(c), the total success rates of E3AC:DC-ED2 and APE2:DC-ED2 throughout training are only slightly lower than those achieved using DC-RL (i.e., E3AC:DC-RL and APE2:DC-RL). According to Table \ref{TABLE 7.5}, this very slight reduction in success rates is mainly reflected in the early training stages.

\begin{figure}[!t]
\centering
\includegraphics[width=2.7in]{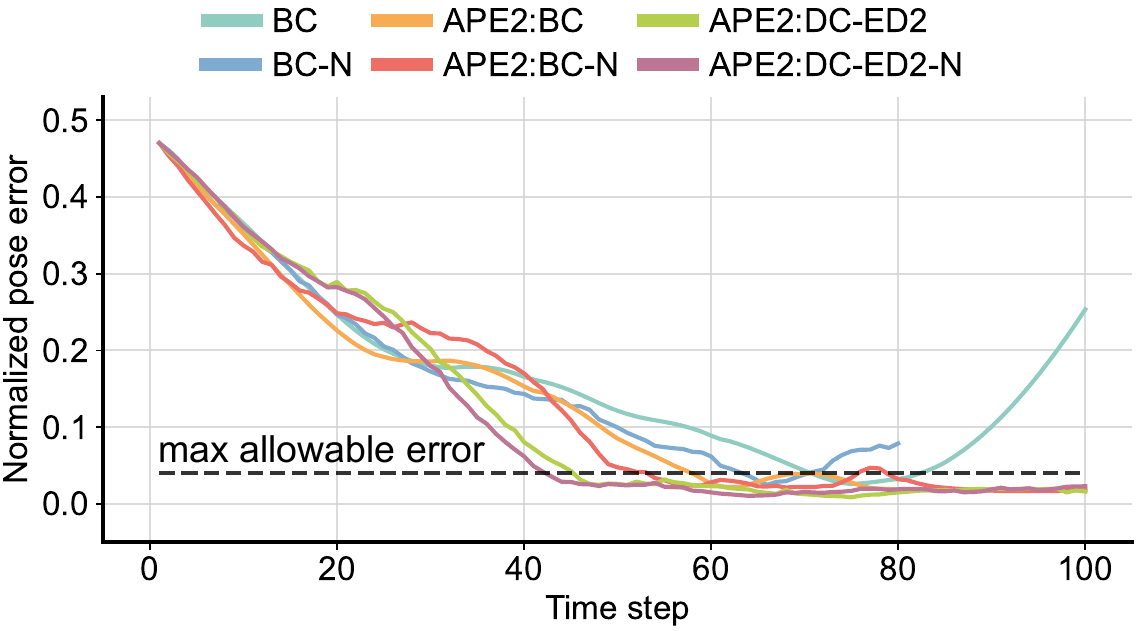}
\caption{Robustness analyses among different uses of the proposed expert data diffusion strategy. The robustness is evaluated from two aspects: the ability to keep the tool center point within allowable errors after reaching the goal pose, and the ability to withstand noise disturbances.}
\label{Fig7.4}
\end{figure}

Moreover, the robustness of well-trained BC, APE2:BC, and APE2:DC-ED2 models based on ED2 has been evaluated from two aspects: the ability to keep the tool center point within allowable errors after reaching the goal pose and the ability to withstand noise disturbances. Fig. \ref{Fig7.4} shows the variation of normalized pose error for each model in motion planning, given a fixed goal pose. Since the diffusion step of ED2 is 80, BC can only plan the motion with fixed 80 steps. Consequently, the pose error of BC increases continuously from the 80th time step onwards, as it cannot maintain the tool center point within allowable errors. Additionally, when Gaussian noise with a variance of 0.4 is added to the action space (see BC-N), a distribution shift phenomenon is observed, with the error of BC-N exceeding the maximum allowable range by the end of the planning task. Compared with BC, the proposed hybrid policy (see APE2:BC) uses only about 60 steps to reduce the pose error to allowable ranges and maintains it within this range for subsequent steps. Furthermore, the hybrid policy shows good robustness against noise disturbances (see APE2:BC-N). 

Comparing APE2:BC and APE2:DC-ED2, it can be inferred that the proposed data compensation mechanism helps DRL learn better motion planning policies. Specifically, the pose error of APE2:DC-ED2 is reduced to within the allowable range in about 50 steps. The subsequent error fluctuation remains minimal and stable, even under noise disturbances (see APE2:DC-ED2-N). In general, pure DRL-based models exhibit greater robustness than BC-based models. Since APE2:DC-ED2 is entirely independent of behavior cloning, its motion planning performance is not constrained by the quality of the provided expert demonstrations. 

Based on the experimental results presented in this section, the proposed ED2 model can effectively generate substantiate expert trajectory data, which can significantly enhance DRL training when utilized with appropriate data integration methods. Moreover, the data compensation mechanism outperforms the hybrid policy learning approach in terms of both training efficiency and trajectory generation quality. Consequently, we adopt the data compensation mechanism as the primary technique for efficient policy learning in the proposed URPlanner.

The results in Fig. \ref{Fig7.4} also demonstrate the capacity of the proposed URPlanner to handle tracking errors effectively. In this work, the trained policy is rolled out in the parameterized environment in a closed-loop feedback manner to generate trajectories (represented as joint configuration sequences). As shown in Fig. \ref{Fig7.4}, even if the actual robot configurations deviate from the originally commanded configurations because of the noise disturbances introduced to the action space at each time step, URPlanner can continue to plan and accomplish the assigned tasks with satisfactory performance. After the trajectory is planned in a closed-loop manner and its safety is verified, the joint configuration sequence is transmitted to the real-world robot for execution in an open-loop manner. In such cases, if the robot fails to reach the commanded configuration at any step due to disturbances, the error is automatically corrected as it moves toward the next joint configuration in the sequence. Moreover, a feedback loop can be implemented in URPlanner to replan subsequent trajectories at a specific frequency. For instance, URPlanner can periodically read the actual joint angles of the robot every few steps to calculate the current state and update all the configurations yet to be tracked. The replanning can be accomplished within a few milliseconds by URPlanner, as demonstrated in the next section.

\subsection{Application Instances}
\label{sec:VII-C}
In this part, the proposed URPlanner is compared with other existing traditional and learning-based motion planners to verify its superiority in trajectory generation. 

\begin{figure}[!t]
\centering
\vspace{0cm} 
\includegraphics[width=0.9\columnwidth]{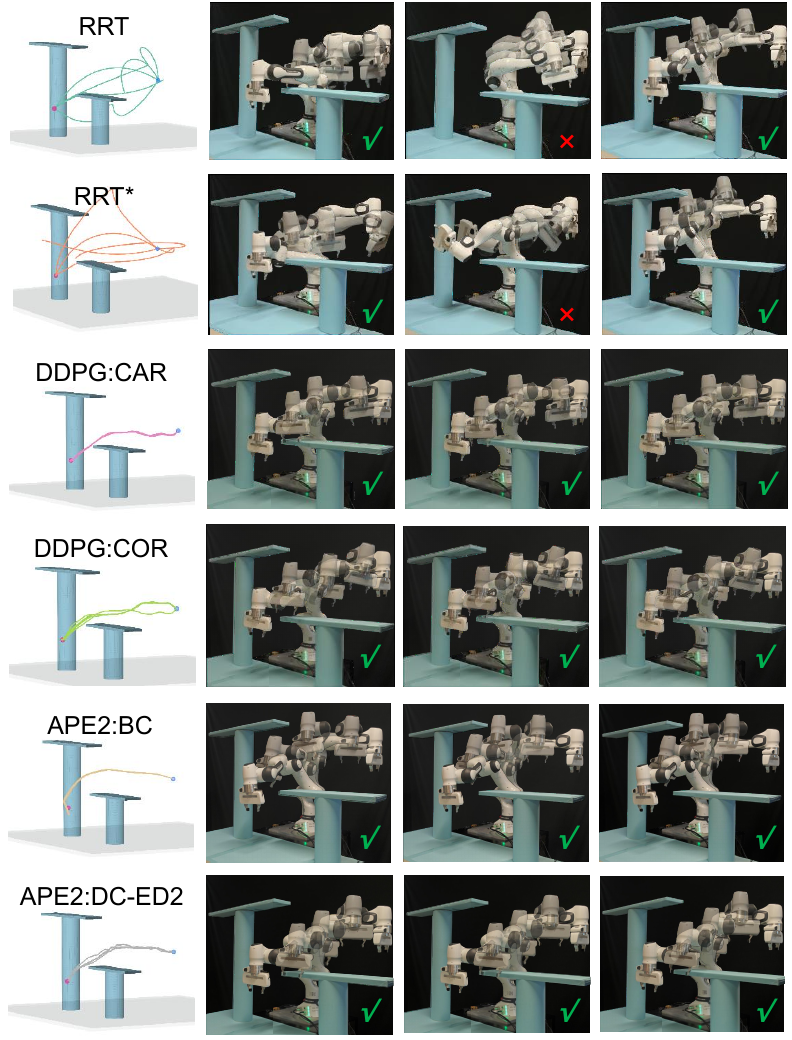}
\vspace{-0.05in}
\caption{Trajectory generation performance of each planner given a fixed goal pose. Column 1 contains all the five trajectories. Columns 2-4 are three selected real-world examples. The green tick and red cross indicate success and failure, respectively.}
\label{Fig7.6}
\end{figure}

\subsubsection{Comparisons among existing planners}
As shown in Fig. \ref{Fig7.6}, we evaluate the quality of trajectories planned by existing planners for a given goal pose. RRT and RRT{$^*$} are selected as representatives of traditional methods, with a maximum planning time of 4 s. DDPG:CAR and DDPG:COR, as proposed in \cite{Fengkang2024}, are powerful learning-based planners. The trajectories for these four methods are generated using a virtual twin system \cite{FengkangASME} established in CoppeliaSim and then applied to real robots, as they are not platform-agnostic. Specifically, for DDPG:CAR and DDPG:COR, the policy is rolled out in the virtual twin system in a closed-loop feedback fashion. The agent observes the current state, computes an action using the trained policy, and transitions to the next state. This process continues iteratively until the goal is achieved. Hence, the generation cost (GC) depends on both the simulation runtime and the inference time of the policy.

In our evaluation of URPlanner, we assess the performance of both APE2:BC and APE2:DC-ED2 to determine which approach offers better results. As platform-agnostic methods, they use the proposed parameterized space to represent the environment with static obstacle layouts, enabling all necessary information for trajectory generation to be computed analytically. Thus, trajectory generation is reduced to a direct computation of states and actions, without requiring a simulated robot platform. Here, the generation cost depends solely on the policy's inference time, significantly lowering the time cost for generating feasible trajectories.

The experiment is repeated 5 times for each planner, and only successful trials are counted when calculating the trajectory length (TL) and the MD between the robot and obstacles during motion. Note the MD data in Table \ref{TABLE 7.7} are obtained using the MD module in CoppeliaSim by verifying the generated trajectories in the virtual twin system.

\begin{table*}
\centering
\vspace{0cm} 
\setlength{\abovecaptionskip}{0cm} 
\setlength{\belowcaptionskip}{0cm} 
\renewcommand{\arraystretch}{1.35}
\caption{Performance Comparisons of Different Motion Planners}
\label{TABLE 7.7}
\setlength{\tabcolsep}{1.2mm}

\begin{threeparttable}
\begin{tabular}{ccccccccccc}

\Xhline{0.7pt}
\specialrule{0em}{1pt}{1pt} 
  \textbf{Category} &
  \textbf{Planner}    &
  \textbf{\makecell{Training\\Cost}}    &
  \textbf{\makecell{Generation\\Cost (s)}}    &
  \textbf{\makecell{Success\\Rate}}    &
  \textbf{\makecell{Trajectory\\Length (m)}}    &
  \textbf{\makecell{Minimum\\Distance (cm)}}    &
  \textbf{Score}    &
  \textbf{IK-free}    &
  \textbf{\makecell{Platform\\Agnostic}}    &
  \textbf{Stability}    

\\
\specialrule{0em}{1pt}{1pt} 
\hline
\specialrule{0em}{1pt}{1pt} 
  \multirow{2}{*}{\textbf{\makecell{Traditional\\Planner}}}                                             &
  RRT                             &
  0                               &  
  $1.937\:_{0.281}^{4.0}$         &
  80.0\%                          &  
  $1.340\:_{1.071}^{1.589}$       &
  $2.62\:_{0.09}^{5.34}$          &
  3.081                           &
  $\times$                        &
  $\times$                        &
  normal         
\\
                                  &
  RRT$^*$                         &
  0                               &  
  $4.0\:_{4.0}^{4.0}$             &
  60.0\%                          &  
  $1.457\:_{1.102}^{1.676}$       &
  $1.68\:_{1.15}^{2.74}$          &
  1.986                           &
  $\times$                        &
  $\times$                        &
  normal       
\\
\specialrule{0em}{1pt}{1pt} 
\hline
\specialrule{0em}{1pt}{1pt} 
  \multirow{4}{*}{\textbf{\makecell{Learning\\Based}}}           &
  DDPG:CAR                        &
  11h44min                        &  
  $6.34\:_{6.24}^{6.44}$          &
  100.0\%                         &  
  $1.153\:_{1.140}^{1.164}$       &
  $5.22\:_{5.06}^{5.41}$          &
  2.802                           &
  \checkmark                      &
  $\times$                        &
  good
\\
                                  &
  DDPG:COR                        &
  9h48min                         &  
  $6.47\:_{6.26}^{6.74}$          &
  100.0\%                         &  
  $1.098\:_{1.080}^{1.140}$       &
  $5.36\:_{5.07}^{6.03}$          &
  3.136                           &
  \checkmark                      &
  $\times$                        &
  good      
\\
                                  &
  APE2:BC                         &
  13min4s                         &  
  $0.212\:_{0.207}^{0.218}$       &
  100.0\%                         &  
  $1.285\:_{1.277}^{1.296}$       &
  $4.92\:_{4.68}^{5.13}$          &
  4.293                           &
  \checkmark                      &
  \checkmark                      &
  good   
\\
                                  &
  APE2:DC-ED2                     &
  \textbf{8min49s}                         &  
  $\textbf{0.060}\:_{0.060}^{0.061}$       &
  \textbf{100.0\%}                         &  
  $\textbf{1.098}\:_{1.089}^{1.108}$       &
  $\textbf{5.47}\:_{5.11}^{5.92}$          &
  \textbf{4.987}                           &
  \checkmark                      &
  \checkmark                      &
  good     
\\
\specialrule{0em}{1pt}{1pt} 
\Xhline{0.7pt}
\\
\end{tabular}
\begin{tablenotes}[para,flushleft]
	\footnotesize
	\vspace{-0.5cm} 
	\item[] \textbf{Annotation:} The data of planning time, trajectory length, and minimum distance between the manipulator and obstacles are in the format of $\rm{mean_{min}^{max}}$. The trajectories are generated in a simulated or parameterized environment to ensure safety, and subsequently sent to the real-world robot in an open-loop fashion. The generation cost refers to the time required by various methods to produce a feasible trajectory for a given task. For non-platform-agnostic methods, this cost includes both the simulation runtime and the policy inference time. In contrast, for our platform-agnostic methods, it depends solely on the policy inference time.
\end{tablenotes}
\end{threeparttable}
\end{table*}

\begin{figure}[!t]
\centering
\includegraphics[width=3in]{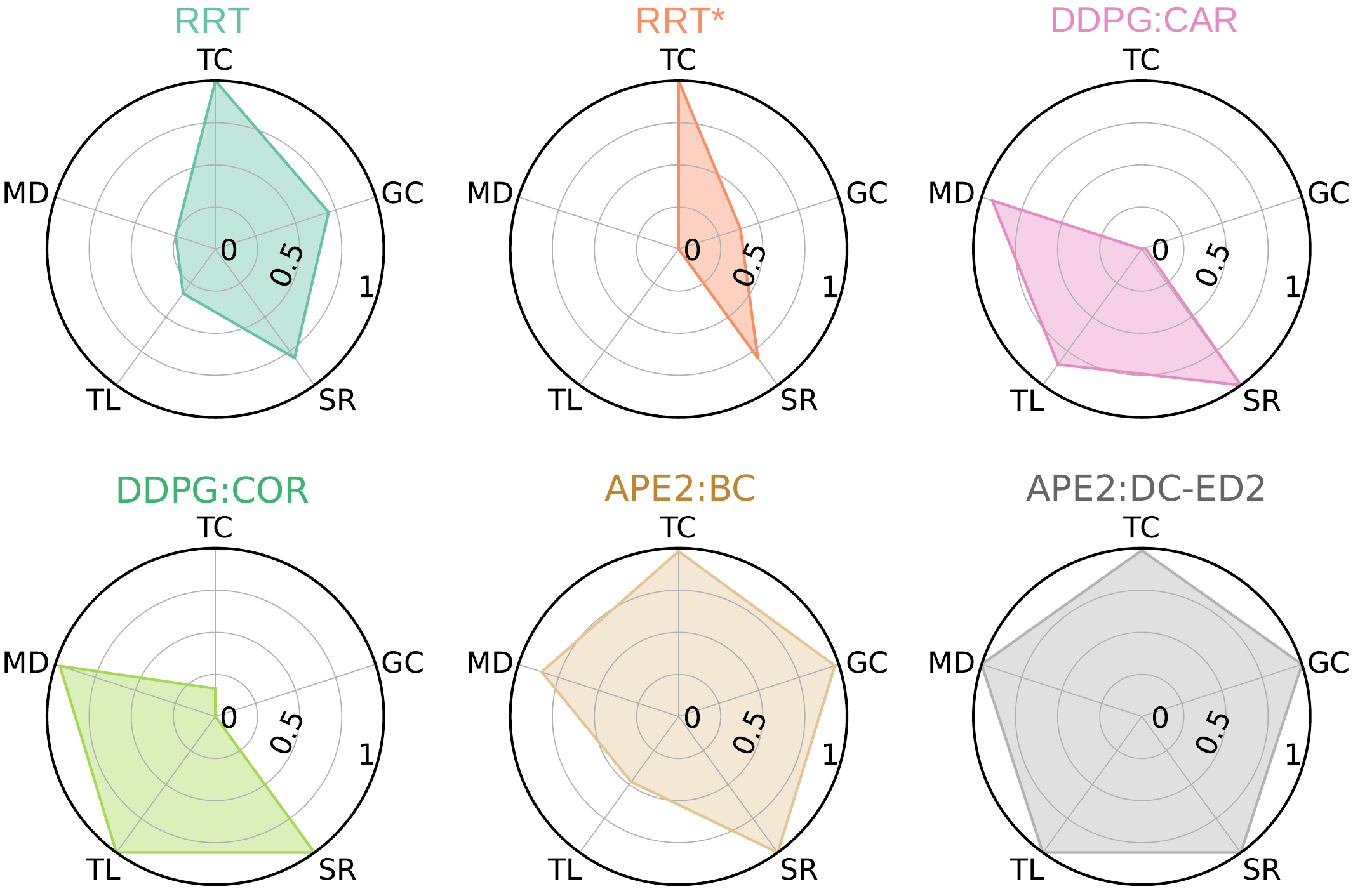}
\caption{Performance evaluation of different motion planners. TC: training cost. GC: generation cost. SR: success rate. TL: trajectory length. MD: minimum distance between the manipulator and obstacle during motion.}
\label{Fig7.5}
\end{figure}

As shown in Table \ref{TABLE 7.7}, the training cost (TC) of RRT and RRT{$^*$} is zero, as they do not need training. However, the performance is highly unstable. For instance, the GC, TL, and MD for RRT vary significantly, ranging from 0.281 s to 4.0 s, 1.071 m to 1.589 m, and 0.09 cm to 5.34 cm, respectively. This instability is attributed to the infinite IK solutions for redundant manipulators and the inherent randomness of the tree-based search process, leading to occasional failures during the five trials. Notably, RRT{$^*$} stabilizes the generation cost at the maximum value of 4 s, attempting to asymptotically find the optimal solution. However, the random process often renders the 4 s planning time insufficient for identifying a solution, resulting in the lowest success rate (SR) of only 60\%. Similar to RRT, the TL and MD of RRT{$^*$} also vary significantly. Moreover, even for the successful trajectories generated by RRT and RRT{$^*$}, the MD is often too small to guarantee safety during motion.


Compared to RRT and RRT{$^*$}, a major drawback of learning-based methods is their high training cost. We train the four planners until their policies converge to a stable state. As shown in Table \ref{TABLE 7.7}, training DDPG:COR and DDPG:CAR takes about 10 h and 12 h, respectively. This lengthy training period is due to the need for interactions with simulated or real-world platforms to acquire necessary environmental information and their reliance on precise MD calculations, as discussed in Section \ref{Section_VII.B1}. For the same reason, both planners require over 6 s to generate a trajectory. In contrast, the APE2:BC and APE2:DC-ED2 models of our URPlanner, as platform-agnostic methods, require only a few minutes for training and a few milliseconds for trajectory generation. This efficiency is achieved using the proposed parameterized space and MD-independent reward. Particularly, the APE2:DC-ED2 model learns a stable policy in less than 9 min, and requires only 0.06 seconds to generate an entire trajectory.

Although the four learning-based motion planners incur certain training costs, they demonstrate good stability and satisfactory trajectory quality. This is because DRL enables the agent to learn an optimal policy to maximize the return (episode reward). As shown in Table \ref{TABLE 7.7}, all of them achieved a 100\% success rate in all the tests. It is also noteworthy that, compared to traditional planners, the average length of the trajectories planned by these models is shorter, while the corresponding minimum distance (between the manipulator and the obstacles throughout the motion process) is larger, indicating superior trajectory quality and safety. Moreover, the trajectory length and minimum distance of the five trajectories planned by each learning-based method exhibit significantly lower variance compared to traditional planners, indicating better stability in motion planning. Additionally, the four planners are designed to be IK-free, allowing them to be applied to arbitrary manipulators without requiring inverse kinematics planning.

By further comparing among the four DRL-based models, it is worth noting that the trajectories planned by APE2:BC show the largest trajectory length of 1.285 m. Additionally, although the MD is satisfactory, it is still slightly smaller than the other models. This is because directly incorporating a non-optimal BC policy in the action ${\bm a}_t$ can also restrict the overall behavior of the hybrid policy, even though it helps accelerate the training of DRL algorithms. In comparison, using the DC mechanism not only brings greater promotion to the training performance (as discussed in Section \ref{sec:VIIB3}, but also ensures the DRL agent can effectively search for and learn an optimal policy. As shown in Table \ref{TABLE 7.7}, the trajectories planned by APE2:DC-ED2 model exhibit the shortest average TL while maintaining the largest average MD.

As described in Section \ref{Sec:III-B}, a safety offset $a_{\rm o}$ is introduced to maintain a safe distance from obstacles throughout the motion. As shown in Table \ref{TABLE 7.7}, with $a_{\rm o}=5$ cm, the trajectories planned by the APE2:DC-ED2 have a TL of ${1.098}\:_{1.089}^{1.108}$ m and a MD of ${5.47}\:_{5.11}^{5.92}$ cm. However, excessively large safety offsets can over-constrain the motion planner. To investigate this trade-off, we analyzed the effect of different $a_{\rm o}$ on both TL and MD. When $a_{\rm o}=0$ cm, the trajectories planned by the APE2:DC-ED2 have a TL of ${1.053}\:_{1.042}^{1.058}$ m and a MD of ${1.58}\:_{1.00}^{1.91}$ cm. When $a_{\rm o}=2.5$ cm, the TL and MD are ${1.069}\:_{1.060}^{1.077}$ m and ${2.75}\:_{1.68}^{3.48}$ cm, respectively. When $a_{\rm o}=8.0$ cm, the planner fails to identify a solution. 

These results indicate that setting the $a_{\rm o}$ too small (e.g., 0 cm) can yield slightly shorter trajectories but compromises safety, as even minor disturbances may lead to collisions during trajectory tracking. Conversely, an overly large $a_{\rm o}$ (e.g., 8.0 cm) may render the task infeasible, as no robot configuration would exist that satisfies the constraints. Our empirical analysis suggests that a safety offset in the range of 2.5-5 cm provides a practical balance between safety and task feasibility. Based on these findings, and consistent with the setup in prior works \cite{FengkangASME}, we adopt a 5 cm safety offset in our framework.

In Fig. \ref{Fig7.5}, we score the overall performance of each motion planner by synthesizing the aforementioned criteria: TC, GC, SR, TL, and MD. The values in Table \ref{TABLE 7.7} are scaled to the range $[0,1]$ using min-max normalization, except for the SR. For all criteria except SR and MD, the score is calculated as (1$-$normalized value). For MD, the normalized value is used directly as the score. Consequently, the APE2:DC-ED2 model of URPlanner achieves the highest score of 4.987, followed by 4.293 for APE2:BC, 3.136 for DDPG:COR, 3.081 for RRT, 2.802 for DDPG:CAR, and 1.986 for RRT{$^*$}.


\begin{table*}
\centering
\vspace{0cm} 
\setlength{\abovecaptionskip}{0cm} 
\setlength{\belowcaptionskip}{0cm} 
\renewcommand{\arraystretch}{1.4}
\caption{Performance of Different Motion Planners with Varied Goal Poses}
\label{TABLE 7.8.2}
\setlength{\tabcolsep}{1.2mm}

\begin{threeparttable}
\begin{tabular}{ccccc|cccc|cccc}

\Xhline{0.7pt}
            &
\multicolumn{4}{c}{\multirow{1}{*}{\textbf{point 1}}}   &
\multicolumn{4}{c}{\multirow{1}{*}{\textbf{point 2}}}   &
\multicolumn{4}{c}{\multirow{1}{*}{\textbf{point 3}}}  
\\
\specialrule{0em}{1pt}{1pt} 
\cmidrule(r){2-5} \cmidrule(r){6-9} \cmidrule(r){10-13}
\specialrule{0em}{1pt}{1pt} 
  \textbf{Planner}    &
  \textbf{\makecell{SR}}    &
  \textbf{\makecell{TL (m)}}    &
  \textbf{\makecell{MD (cm)}}    & 
  \textbf{\makecell{Score}}    &
   \textbf{\makecell{SR}}    &
  \textbf{\makecell{TL (m)}}    &
  \textbf{\makecell{MD (cm)}}    & 
  \textbf{\makecell{Score}}    &
   \textbf{\makecell{SR}}    &
  \textbf{\makecell{TL (m)}}    &
  \textbf{\makecell{MD (cm)}}   &
  \textbf{\makecell{Score}}   

\\
\specialrule{0em}{1pt}{1pt} 
\hline
\specialrule{0em}{1pt}{1pt} 
  RRT                                      &
  100.0\%                           &
  $1.147\:_{0.956}^{1.391}$       &  
  $2.76\:_{0.55}^{4.33}$         &
  1.625                               &
  60.0\%                          &  
  $0.982\:_{0.971}^{0.999}$       &
  $4.35\:_{0.41}^{7.74}$          &
  1.582                               &
  60.0\%                        &
  $0.948\:_{0.835}^{1.131}$                        &
  $1.55\:_{0.08}^{2.70}$          &
  1.542
\\
  RRT$^*$                                &
  60.0\%                         &
  $1.180\:_{1.162}^{1.215}$       &  
  $1.99\:_{0.80}^{4.15}$             &
  0.772                               &
  60.0\%                          &  
  $1.206\:_{0.887}^{1.461}$       &
  $4.58\:_{2.77}^{6.83}$          &
  0.653                               &
  80.0\%                        &
  $1.007\:_{0.891}^{1.144}$       &
  $4.77\:_{0.55}^{7.76}$          &
  1.762                 
\\
  DDPG:CAR                        &
  100.0\%                        &
  $1.196\:_{1.182}^{1.203}$       &  
  $7.81\:_{7.31}^{8.36}$             &
  1.745                               &
  100.0\%                          &  
  $1.057\:_{1.049}^{1.063}$       &
  $7.68\:_{7.44}^{7.96}$          &
  2.414                               &
  100.0\%                        &
  $1.046\:_{1.019}^{1.119}$       &
  $5.74\:_{4.46}^{6.09}$          &
  1.763
\\
  DDPG:COR                                &
  100.0\%                        &
  $1.154\:_{1.145}^{1.164}$       &  
  $6.78\:_{6.57}^{7.05}$             &
  2.065                               &
  100.0\%                          &  
  $1.067\:_{1.060}^{1.077}$       &
  $6.08\:_{5.95}^{6.15}$          &
  2.005                               &
  100.0\%                        &
  $0.965\:_{0.958}^{0.972}$       &
  $4.70\:_{4.21}^{5.35}$          &
  2.353      
\\
  APE2:BC                                &
  100.0\%                        &
  $1.144\:_{1.131}^{1.180}$       &  
  $8.46\:_{8.14}^{8.69}$             &
  2.388                               &
  100.0\%                          &  
  $1.071\:_{1.063}^{1.077}$       &
  $8.02\:_{7.69}^{8.66}$          &
  2.430                               &
  100.0\%                        &
  $1.040\:_{1.005}^{1.076}$       &
  $6.85\:_{6.33}^{7.34}$          &
  2.023   
\\
  APE2:DC-ED2                                &
  \textbf{100.0\%}                     &
  $\textbf{1.103}\:_{1.096}^{1.108}$                         &  
  $\textbf{9.80}\:_{9.55}^{10.0}$       &
  \textbf{3.000}                               &
  \textbf{100.0\%}                         &  
  $\textbf{0.978}\:_{0.951}^{0.994}$       &
  $\textbf{8.73}\:_{8.57}^{9.05}$          &
  \textbf{3.000}                               &
  \textbf{100.0\%}                         &  
  $\textbf{0.942}\:_{0.936}^{0.949}$       &
  $\textbf{7.04}\:_{6.35}^{7.42}$          &
  \textbf{3.000}     
\\
\specialrule{0em}{1pt}{1pt} 
\Xhline{0.7pt}
\\
\end{tabular}
\begin{tablenotes}[para,flushleft]
	\footnotesize
	\vspace{-0.5cm} 
    \item[] \textbf{Annotation:} The data of trajectory length as well as minimum distance between the manipulator and obstacles are in the format of $\rm{mean_{min}^{max}}$. 
\end{tablenotes}
\end{threeparttable}
\end{table*}

Without loss of generality, the performances of the six planners are further evaluated by comparing the quality of the trajectories planned for several other goal poses within the designated target area. The distance between two consecutive points is 10 cm. Similar to the experiments presented in Table \ref{TABLE 7.7}, five trajectories are generated by each planner for each goal pose, with only successful trials included in the calculations of trajectory length and minimum distance. Table \ref{TABLE 7.8.2} presents the success rate of trajectory generation, trajectory length, and minimum distance for each planner. It can be observed that RRT and RRT{$^*$} remain highly unstable, with significant variability in the quality of the generated trajectories across the five trials. In contrast, the other four DRL-based planners exhibit good stability and satisfactory trajectory quality, with the trajectories almost perfectly overlapping. The trajectory generation performance is scored based on these three aspects using the calculation method introduced in Fig. \ref{Fig7.5}. The experimental results demonstrate that the APE2:DC-ED2 model of our URPlanner consistently outperforms the other planners in the three listed criteria. It is worth noting that the trajectory generation policies of DRL-based planners are highly similar for goal poses in close proximity to each other. Based on the experimental results in this section, it can be concluded that the APE2:DC-ED2 model of URPlanner achieves the best overall performance among the six planners.

\subsubsection{Trajectory replanning}

\begin{figure}[!t]
\centering
\setlength{\abovecaptionskip}{-0.1cm} 
\includegraphics[width=2.3in]{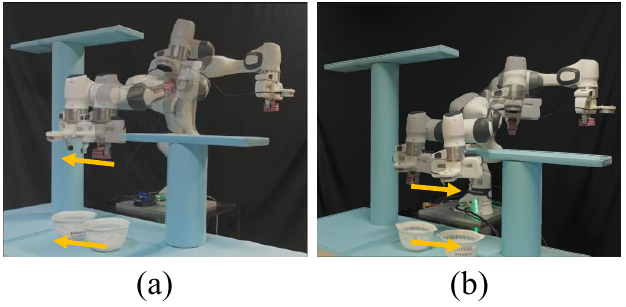}
\caption{Examples of trajectory replanning for changed goal poses. The robot initially executes a preplanned trajectory to transport a grasped object above a static basket. Subsequently, the basket moves in the direction indicated by the yellow arrow. URPlanner replans the subsequent motions, enabling the robot to track the moving basket.}
\label{Fig7.7}
\end{figure}

As discussed in Sections \ref{Section_VII.B} and \ref{sec:VII-C}, the proposed platform-agnostic method, URPlanner, enables all necessary information for trajectory generation to be computed analytically. Due to this advantage, URPlanner can preplan the entire trajectory (represented as joint configuration sequences) efficiently in the parameterized space before execution on a simulated or real-world platform. Moreover, it can replan and update all untracked configuration sequences in the parameterized space concurrently with trajectory execution. As indicated in Table \ref{TABLE 7.7}, the planning can be implemented within a few milliseconds. In Fig. \ref{Fig7.7}, we present two examples of trajectory replanning to accommodate new goal positions. Initially, the Franka robot executes a trajectory preplanned by URPlanner to transport a grasped object above a static basket. Subsequently, in the two examples, the basket moves in the direction indicated by the yellow arrow. URPlanner replans the subsequent motions in the parameterized space, enabling the robot to track the basket's movement and successfully drop the object into it. Notably, even as the goal pose eventually exceeds the trained target area, URPlanner demonstrates robustness to these untrained environmental states.

\section{Discussion}
\subsection{Universality of URPlanner}

\begin{figure*}[!t]
\centering
\includegraphics[width=6.2in]{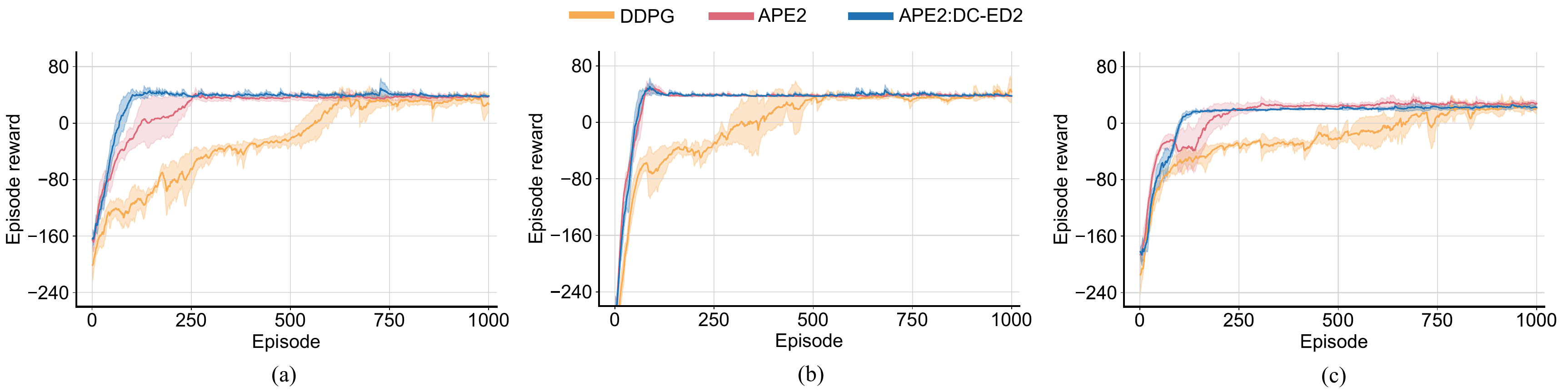}
\caption{Variation of episode rewards during training for DDPG, APE, and APE2:DC-ED2 algorithms in different scenarios. The shaded region represents the standard deviation of the average evaluation over four random seeds. (a) Scenario 1: training a KUKA robot. (b) Scenario 2: training a Franka robot in a simple environment. (c) Scenario 3: training a Franka robot in a complex environment.}
\label{Fig8.2}
\end{figure*}

In this section, we demonstrate the universality and effectiveness of URPlanner across various task scenarios, as illustrated in Fig. \ref{Fig8.1}. These scenarios involve different manipulators and environments. For each scenario, we compare the training performance of DDPG, APE2, and APE2:DC-ED2 algorithms. The training success rates are summarized in Table \ref{TABLE 8.1}, and the variation of episode rewards during training is depicted in Fig. \ref{Fig8.2}. The benchmark algorithm, DDPG, exhibits much slower policy convergence compared to the proposed APE2 and APE2:DC-ED2 algorithms across all scenarios. In contrast, APE2 achieves stable policy learning within approximately 250 episodes in more complex environments, such as Scenarios 1 and 3. Scenario 2, being significantly simpler, allows the APE2 algorithms to learn a stable policy in only about 80 episodes. A comprehensive comparison of training success rates during the early stages (1st-500th episodes) reveals that the proposed APE2 algorithm demonstrates substantial improvements over DDPG.

Despite the powerful learning ability APE exhibits in handling environments with different levels of complexity, its performance can be further enhanced by the proposed expert data diffusion approach. Experimental results indicate that APE2:DC-ED2 achieves the best training performance across all scenarios, particularly during the first 250 episodes, where it learns a stable policy within approximately 70 episodes for Scenarios 1 and 3. Notably, the improvement in training success rate brought by APE2:DC-ED2 in Scenario 2 is limited, as APE2 already performs exceptionally well in this simple environment. In practice, expert data are primarily required to address highly complex scenarios.

These findings demonstrate that URPlanner is a universal paradigm capable of supporting different manipulators and environments. The proposed techniques, including collision avoidance, policy exploration and evaluation, as well as expert data diffusion and utilization, are effective across diverse scenarios. Furthermore, Fig. \ref{Fig8.1} highlights the platform-agnostic nature of URPlanner, as policies trained in the parameterized space can be directly applied to both simulated and real-world platforms without fine-tuning.

\begin{table*}
\centering
\vspace{0cm} 
\setlength{\abovecaptionskip}{0cm} 
\setlength{\belowcaptionskip}{0cm} 
\renewcommand{\arraystretch}{1.4}
\caption{Training Success Rate (\%) per 250 Episodes of Algorithms in Different Scenarios}
\label{TABLE 8.1}
\setlength{\tabcolsep}{1.25mm}

\begin{threeparttable}
\begin{tabular}{ccccc|cccc|cccc}
\Xhline{0.7pt}
  &
  \multicolumn{4}{c}{\multirow{1.4}{*}{\textbf{Scenario 1}}}  &
  \multicolumn{4}{c}{\multirow{1.4}{*}{\textbf{Scenario 2}}}  &
  \multicolumn{4}{c}{\multirow{1.4}{*}{\textbf{Scenario 3}}}
\\

\specialrule{0em}{1pt}{1pt} 
\cmidrule(r){2-5} \cmidrule(r){6-9} \cmidrule(r){10-13}
\specialrule{0em}{1pt}{1pt} 
  &
  1-250   &  
  251-500 &
  501-750 &
  751-1000&
  1-250   &  
  251-500 &
  501-750 &
  751-1000&
  1-250   &  
  251-500 &
  501-750 &
  751-1000
\\
\specialrule{0em}{1pt}{1pt} 
\hline
\specialrule{0em}{1pt}{1pt} 
  DDPG           &
  $0.0\:_{0.0}^{0.0}$          &
  $0.0\:_{0.0}^{0.0}$         &  
  $54.4\:_{37.2}^{62.0}$        &
  $98.0\:_{94.4}^{99.6}$       
  &
  $0.0\:_{0.0}^{0.0}$          &
  $54.3\:_{14.8}^{81.6}$         &  
  $98.8\:_{97.2}^{99.6}$        &
  $98.6\:_{97.2}^{99.2}$        
  &
  $0.0\:_{0.0}^{0.0}$          &
  $4.5\:_{0.0}^{18.0}$         &  
  $53.4\:_{27.6}^{95.6}$        &
  $95.9\:_{93.6}^{98.0}$
\\
  APE2        &
  $31.2\:_{4.0}^{58.0}$         &
  $99.4\:_{98.4}^{100}$        &  
  $99.8\:_{99.6}^{100}$         &
  $100\:_{100}^{100}$
  &
  $72.4\:_{67.2}^{78.8}$          &
  $100\:_{100}^{100}$         &  
  $99.9\:_{99.6}^{100}$        &
  $99.8\:_{99.6}^{100}$
  &
  $29.9\:_{11.6}^{38.4}$          &
  $100\:_{100}^{100}$         &  
  $98.8\:_{96.4}^{99.6}$        &
  $99.7\:_{99.2}^{100}$
\\
  APE2:DC-ED2    &
  $69.3\:_{63.6}^{79.2}$         &
  $100\:_{100}^{100}$         &  
  $99.4\:_{98.8}^{99.6}$         &
  $99.8\:_{99.2}^{100}$
  &
  $78.3\:_{74.8}^{81.6}$          &
  $99.9\:_{99.6}^{100}$         &  
  $98.8\:_{98.4}^{99.6}$        &
  $99.9\:_{99.6}^{100}$
  &
  $66.5\:_{65.2}^{69.2}$          &
  $100\:_{100}^{100}$         &  
  $99.1\:_{97.2}^{100}$        &
  $99.9\:_{99.6}^{100}$
\\
\specialrule{0em}{1pt}{1pt} 
\Xhline{0.7pt}
\end{tabular}
\begin{tablenotes}[para,flushleft]
	\footnotesize
	\textbf{Annotation:} The data are in the format of $\rm{mean_{min}^{max}}$. In each scenario, all the algorithms achieve a 100\% testing success rate based on 50 trials. 
\end{tablenotes}
\end{threeparttable}
\end{table*}

\begin{figure}[!t]
\centering
\setlength{\belowcaptionskip}{-0.3cm}
\includegraphics[width=0.75\columnwidth]{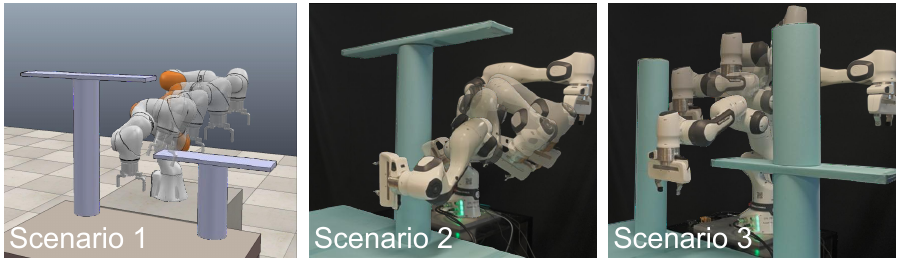}
\caption{Universality of URPlanner across different manipulators and environments. Scenario 1: training a KUKA robot. Scenario 2: training a Franka robot in a simple environment. Scenario 3: training a Franka robot in a complex environment.}
\label{Fig8.1}
\end{figure}

\subsection{Generalization across different scenarios}

The generalization ability is a common challenge in DRL, as optimal policies often vary across different environments. Typically, if the new environment closely resembles the training environment, such as the experiments in Table \ref{TABLE 7.4.2}, the model may generalize reasonably well. However, in cases where the new environment differs significantly, such as featuring entirely new obstacles, manipulators, or target areas, the model may struggle to perform effectively without retraining. Studies including \cite{2019Feedback,FengkangASME,Fengkang2024} emphasize that retraining is often required when applying a learned policy to new task scenarios.

In this work, we abstract obstacles as one or multiple bounding boxes in the proposed parameterized space. This abstraction ensures that the trained motion planning policy is universally applicable to all objects that can be enclosed by the bounding box used during training. While training a single DRL policy to handle all possible task scenarios is inherently challenging, the introduction of a policy library enables our method to generalize effectively across various scenarios.

\begin{figure}[!t]
\centering
\setlength{\belowcaptionskip}{-0.3cm}
\includegraphics[width=2.4in]{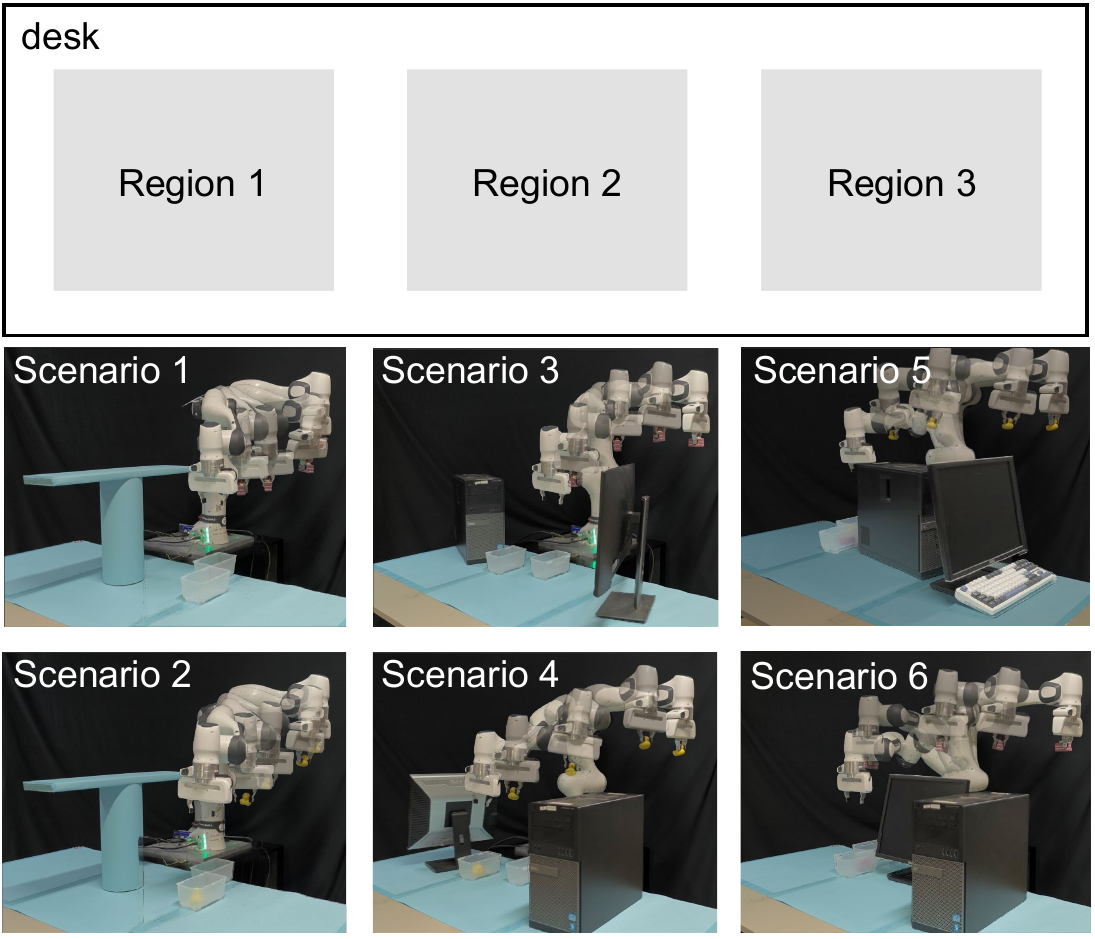}
\caption{Examples of enhancing generalization by developing a policy library. Six new scenarios involving random obstacle and target layouts are given, and the agent can retrieve an appropriate sub-policy from the library to generate collision-free motions.}
\label{Fig8.3}
\end{figure}

As illustrated by the simple example in Fig. \ref{Fig8.3}, the workspace on a desk is divided into three Regions: 1, 2, and 3. Three sub-policies, each tailored to a respective sub-workspace, are trained. To train these policies, each time, one region is defined as a sub-workspace, with the remaining two regions treated as obstacle spaces with a certain height. These sub-policies, represented by network parameters, are stored in the policy library along with the spatial occupancy data of the target and obstacle regions.

\begin{figure}[!t]
\centering
\vspace{-0.1cm} 
\setlength{\belowcaptionskip}{-0.3cm} 
\includegraphics[width=\columnwidth]{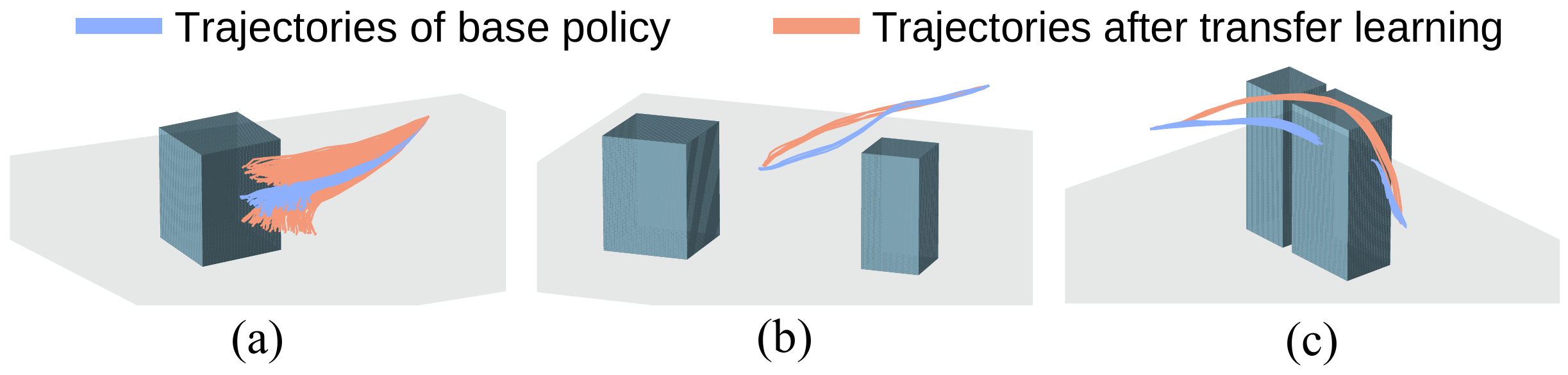}
\caption{Examples of leveraging transfer learning for efficient policy adaptation to new environments. (a) Policy adaptation for new target areas. (b) Policy adjustment for smaller obstacles. (c) Policy adaptation for larger obstacles.}
\label{Fig8.4}
\end{figure}

The combination of these three sub-policies enables coverage of a broad range of obstacle and target layouts, provided that the target is not in the same region as the obstacles. As illustrated in Fig. \ref{Fig8.3}, six scenarios with random goal positions and/or obstacles are established. In Scenarios 1 and 2, the two boxes are located in Region 1, where the obstacles in Region 2 have minimal influence on the robot's motion to the goals. In Scenarios 3 and 4, the boxes are located in Region 2, while the two obstacles are positioned randomly in Regions 1 and 3, respectively. In Scenarios 5 and 6, the robot needs to navigate across two consecutive obstacle regions to reach the goals in Region 3. These examples demonstrate that the agent can flexibly retrieve an appropriate sub-policy from the library to generate collision-free motions, even when faced with new scenarios involving random obstacle and target configurations.

If no feasible sub-policy exists in the library, i.e., an obstacle and target configuration not covered by the existing policies, a new sub-policy is automatically trained and added to the library. The cost effectiveness of the URPlanner proposed in this work enables this training process to be completed within a few minutes, compared to the hours required by conventional DRL-based planners. As the policy library expands with additional sub-policies, the generalization capability of our approach continuously improves. 

Additionally, the training cost for a new sub-policy can be further reduced by leveraging transfer learning \cite{Fengkang2024}. For example, Fig. \ref{Fig8.4}(a) illustrates the policy adaptation for new target areas where the original target area is raised or lowered by 10 cm. The new policies are trained within about 150 s by transferring the base policy. In Fig. \ref{Fig8.4}(c), the obstacles are stretched, causing the base policy unable to generate collision-free trajectories. Within about 75 s of transfer learning, the base policy is adjusted to stably generate collision-free trajectories for reaching random targets in Region 3.

While the policy library effectively improves generalization, it may compromise the quality of the generated trajectories. This is particularly evident when the bounding box used for training the sub-policy is significantly larger than the obstacles in new scenarios, leading to unnecessary safety margins and inefficiencies in reaching the target. In such situations, transfer learning can be employed to refine the trajectory generated by the sub-policy. Fig. \ref{Fig8.4}(b) illustrates an example where transfer learning refines the base policy for reaching random targets in region 2, reducing the average trajectory length by 4.6 cm and increasing the minimum distance to obstacles by 4.3 cm.

\subsection{Strategy for providing demonstrations to ED2}

\begin{figure}[!t]
\centering
\vspace{-0.3cm} 
\setlength{\belowcaptionskip}{-0.3cm} 
\includegraphics[width=3.4in]{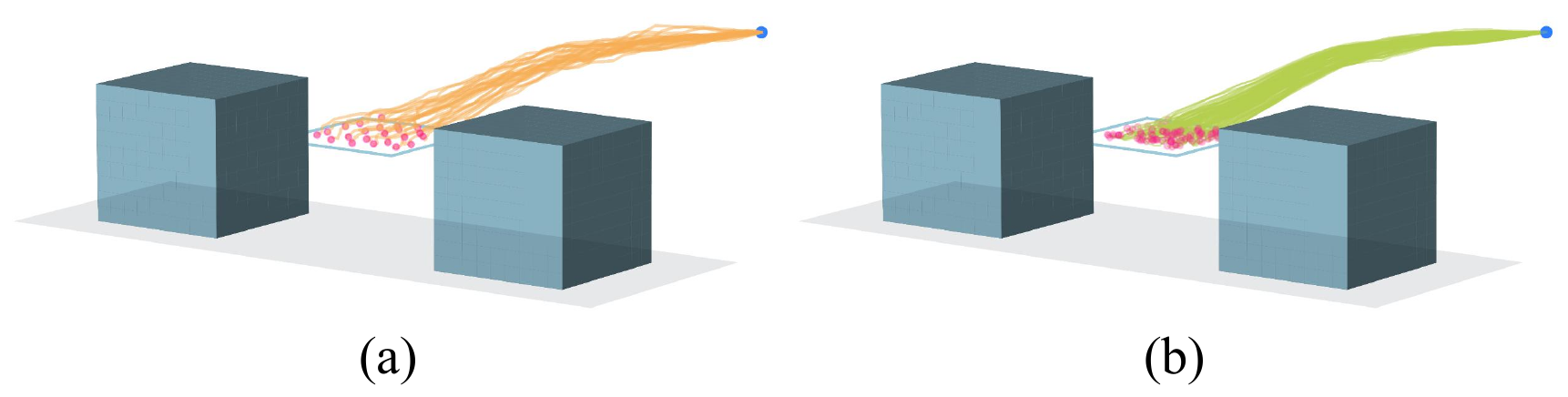}
\caption{Example of demonstration provision and novel trajectory generation by ED2. The designated target area is uniformly divided into 25 sub-areas, with demonstrations provided for each sub-area. (a) Twenty-five provided demonstrations. (b) One hundred novel trajectories generated by ED2.}
\label{Fig8.7}
\end{figure}

\begin{table}
\centering
\vspace{0cm} 
\setlength{\abovecaptionskip}{0cm} 
\setlength{\belowcaptionskip}{0cm} 
\renewcommand{\arraystretch}{1.4}
\caption{Training Performance of the APE2:DC-ED2 Algorithm with Varying Numbers of Expert Demonstrations}
\label{TABLE 8.2}
\setlength{\tabcolsep}{1.7mm}

\begin{threeparttable}
\begin{tabular}{cccccc}
\Xhline{0.7pt}
\multirow{2.5}{*}{\textbf{\makecell[c]{Algo-\\rithm}}}     &
\multirow{2.5}{*}{\textbf{\makecell[c]{Number\\of demo-\\strations}}}
  &
  \multicolumn{4}{c}{\multirow{1.2}{*}{\textbf{Success Rate Per 250 Episodes (\%)}}}
\\

\cmidrule(r){3-6}
  &
  &
  1-250   &  
  251-500 &
  501-750 &
  751-1000
\\
\specialrule{0em}{1pt}{1pt} 
\hline
\specialrule{0em}{1pt}{1pt} 
  APE2           &
  $-$            &
  $7.8\:_{0.4}^{25.2}$         &
  $86.9\:_{75.2}^{98.8}$        &  
  $100\:_{100}^{100}$        &
  $99.5\:_{99.2}^{100}$         
\\
\specialrule{0em}{1pt}{1pt} 
\hline
\specialrule{0em}{1pt}{1pt} 
  \multirow{4}{*}{\makecell[c]{APE2 with\\DC-ED2}}        &
  1   &
  $43.2\:_{36.8}^{48.0}$         &
  $99.9\:_{99.6}^{100}$        &  
  $100\:_{100}^{100}$         &
  $99.7\:_{99.2}^{100}$         
\\
      &
  5   &
  $45.6\:_{43.2}^{48.8}$         &
  $100\:_{100}^{100}$        &  
  $100\:_{100}^{100}$         &
  $99.7\:_{99.6}^{100}$         
\\
      &
  10  &
  $45.8\:_{36.4}^{53.6}$         &
  $100\:_{100}^{100}$        &  
  $99.9\:_{99.6}^{100}$         &
  $99.7\:_{99.2}^{100}$         
\\
      &
  20  &
  $47.1\:_{41.6}^{50.4}$         &
  $99.8\:_{99.6}^{100}$        &  
  $99.9\:_{99.6}^{100}$         &
  $100\:_{100}^{100}$         
\\
\specialrule{0em}{1pt}{1pt} 
\Xhline{0.7pt}
\end{tabular}
\begin{tablenotes}[para,flushleft]
	\footnotesize
	\textbf{Annotation:} The data are in the format of $\rm{mean_{min}^{max}}$ over four random seeds. All the algorithms achieve a 100\% testing success rate based on 50 trials. 
\end{tablenotes}
\end{threeparttable}
\end{table}

As discussed in Section \ref{sec:VII-C}, the trajectory generation policies of DRL-based planners are highly similar for goal poses in close proximity to each other. Thus, a designated target area can be uniformly divided into several sub-areas, with demonstrations provided for each sub-area. For example, in Fig. \ref{Fig8.7}(a), one demonstration is provided for each of the 25 sub-areas. Fig. \ref{Fig8.7}(b) shows 100 novel trajectories generated by the proposed ED2 model, trained using the 25 demonstrations shown in Fig. \ref{Fig8.7}(a). These results demonstrate that the ED2 model is capable of generating diverse novel trajectories from a limited number of expert demonstrations. 

As shown in Table \ref{TABLE 8.2}, by setting the number of demonstrations provided to each sub-area as 1, 5, 10, and 20, respectively, four ED2 models are trained accordingly. Subsequently, four expert memories are generated based on these trained ED2 models. These expert memories are used by the data compensation mechanism, referred to as DC-ED2, to evaluate their impact on facilitating the training of the APE2 algorithm. The results demonstrate that the APE2:DC-ED2 algorithm with ED2 trained on a single demonstration per sub-area achieves a success rate of 43.2\% in the first 250 training episodes. This performance is only marginally lower than the success rates achieved by the APE2:DC-ED2 using ED2 models trained on 5, 10, and 20 demonstrations per sub-area, respectively. These results demonstrate that ED2, trained with a limited number of demonstrations, is capable of generating high-quality data to accelerate the policy convergence of DRL.

\subsection{Extension of URPlanner}

As discussed in Section \ref{Sec:III-B}, URPlanner adopts a clearance-based motion planning strategy by expanding the bounding boxes that represent obstacles. This expansion ensures that the robot maintains a safe distance from potential collisions. However, this conservative approach may limit the planner's ability to navigate through environments with narrow passages or tightly clustered obstacles.

\begin{figure}[!t]
\centering
\includegraphics[width=0.8\columnwidth]{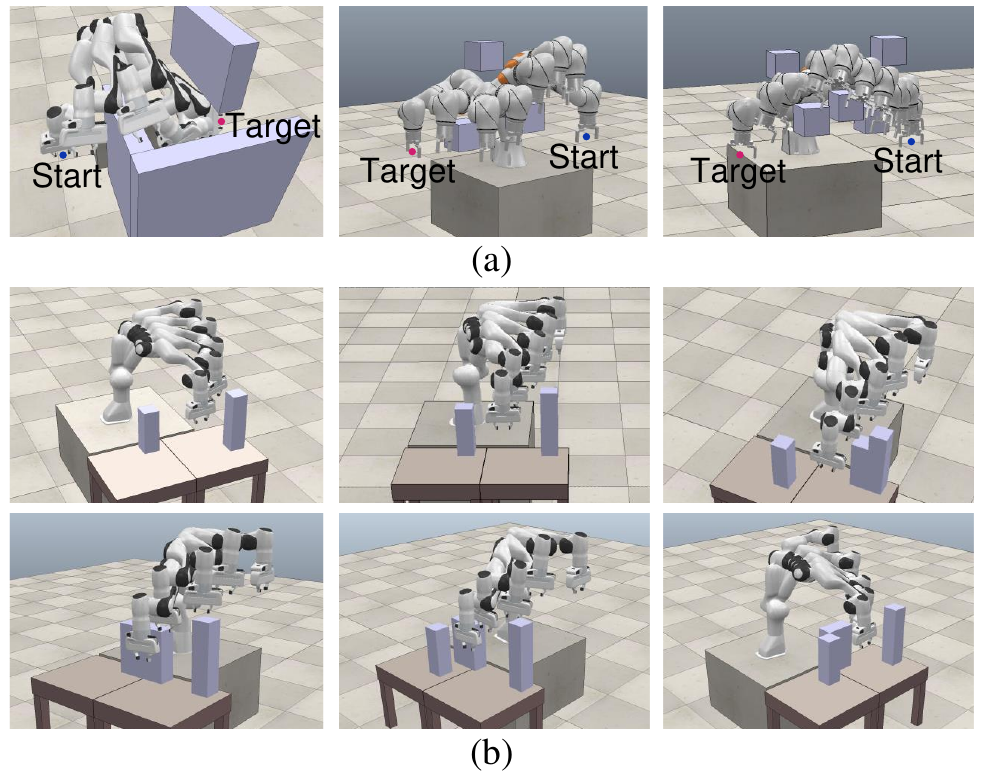}
\caption{Extensions of URPlanner. (a) Approximating the robot's link with multiple line segments eliminates the need to expand obstacle representations and enhances URPlanner's planning ability in narrow spaces. (b) URPlanner can be extended to handle environments with randomized obstacle configurations by incorporating obstacle information as a 3D occupancy grid matrix in the state representation.}
\label{Fig8.8}
\end{figure}

URPlanner can be easily extended to overcome this limitation. Specifically, the surface of each cylinder that represents the robot's link can be approximated by multiple line segments, similar to the method in \cite{Mu2014}. The computation of collision detection and the UOAR remains unchanged. This approach eliminates the need to expand obstacle representations and avoids reducing the cylindrical model to a single central line segment. Although this extension slightly increases the reward computation time, it significantly improves URPlanner's planning ability in constrained environments. As demonstrated in Fig. \ref{Fig8.8}(a), the extended URPlanner effectively guides both KUKA and Franka manipulators to execute smooth, collision-free curved trajectories, successfully navigating narrow spaces between ground-level and elevated obstacles.

Moreover, URPlanner can be extended to handle environments with random obstacle layouts. In this work, we assume static obstacle layouts and only the target positions vary randomly across episodes. In this context, the state ${\bm s}_t$ defined in Eq. \eqref{eq02} is sufficient. This state setup is consistent with many prior works in DRL-based motion planning, such as \cite{Fengkang2024,FengkangTII,FengkangASME}.

To handle environments with random obstacle layouts, obstacle information must be incorporated into the state variable. A straightforward yet limited solution adopted in prior works \cite{CHEN202264,Yang2022,ZZJ2024} is to include the positions of the obstacles. This method works when the number and shape of obstacles are constant, but it becomes impractical in highly variable settings.

To enable scalability, we propose using a 3D occupancy grid with a fixed resolution to represent the spatial configuration of obstacles. This representation allows the state variable to remain fixed in dimension while encoding arbitrary obstacle shapes and distributions. As shown in Fig. \ref{Fig8.8}(b), by incorporating the occupancy grid into ${\bm s}_t$, URPlanner is capable of handling environments with randomized obstacle configurations. The robot successfully learns to navigate around obstacles whose number, size, and shape change across episodes. In future work, we plan to further optimize this approach.

\section{Conclusion}

This article presents a universal paradigm for collision-free robotic motion planning named URPlanner, which is platform-agnostic and highly cost-effective in both training and deployment. This efficiency is attributed to integrating the proposed parameterized task space, minimum distance-independent UOAR, powerful APE2 algorithm, and ED2 model with data compensation mechanism. For a given environment, URPlanner can be trained within a few minutes. The experimental results have also shown that the trained model can plan IK-free, collision-free, and near-optimal motion within a few milliseconds. In contrast, other methods may take up to several seconds or longer. Moreover, our URPlanner demonstrates good robustness and adaptability across various task scenes, platforms, and manipulators, and generalizes well to untrained environmental states.

One limitation of our work is that the approximation of obstacles and manipulator links may lead to slight underutilization of free space, as some collision-free regions are conservatively treated as unsafe. However, this simplification provides significant gains in computational efficiency. In practice, these regions can serve as parts of the safety margins, thereby improving clearance during motion planning. For future work, we think it would be interesting to explore the application of the proposed URPlanner in dynamic environments with moving obstacles. Particularly, a human partner may also be modeled as a moving obstacle so that the approach can also be used in human-robot collaboration scenarios. 

\bibliography{bib_IEEE}

\vspace{-0.1in}
\begin{IEEEbiography}[{\includegraphics[width=1in,height=1.25in,clip,keepaspectratio]{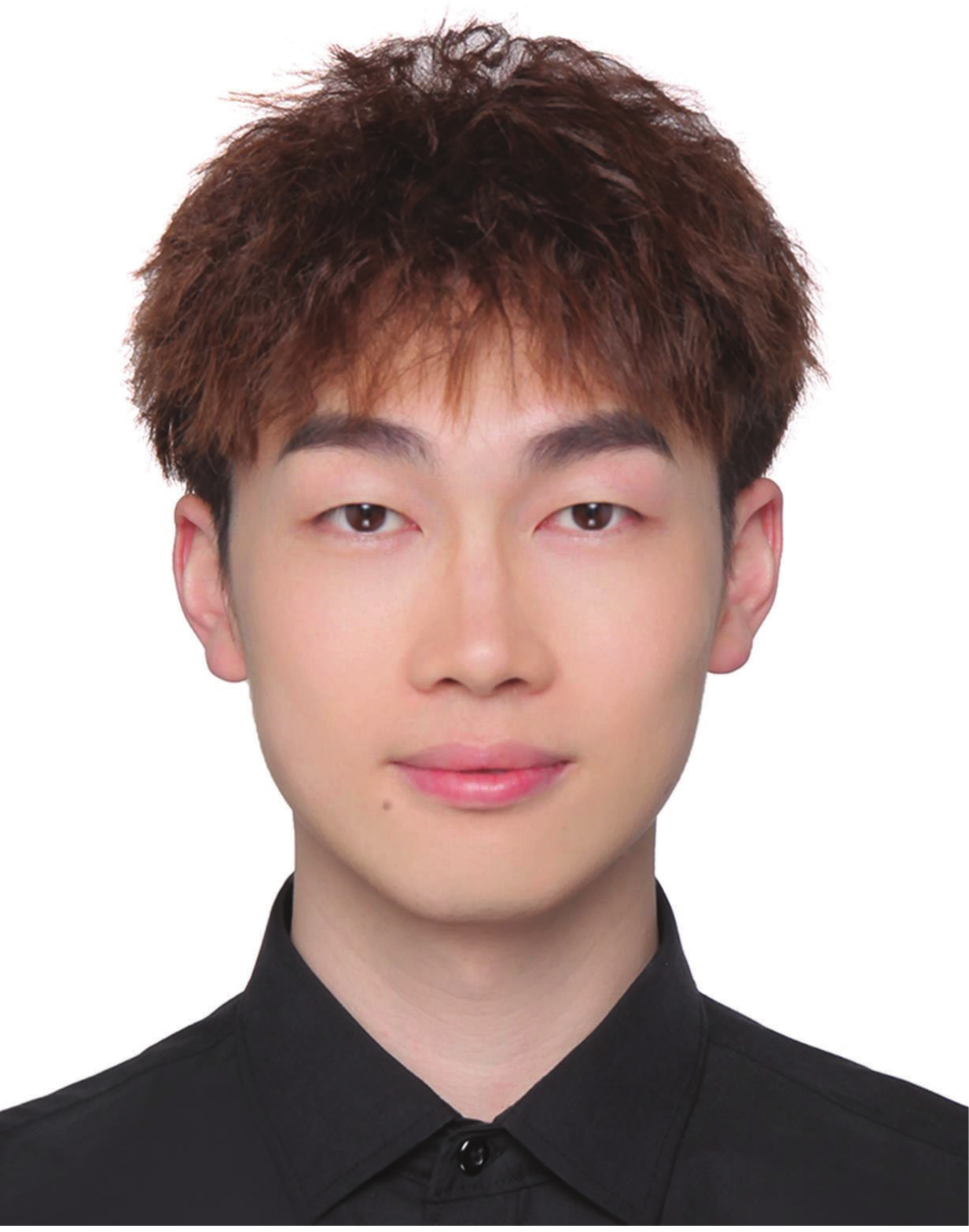}}]{Fengkang Ying} (Graduate Student Member, IEEE)
received the M.E. degree in Electrical Engineering from Donghua University, Shanghai, China, in 2022. He is currently working towards the Ph.D. degree in the Integrative Sciences and Engineering Programme (ISEP), NUS Graduate School, National University of Singapore (NUS), Singapore.

Since August 2023, he has been with the ISEP and NUS Advanced Robotics Centre. His current research interests include robotics, task and motion planning, and reinforcement learning.
\end{IEEEbiography}

\vspace{-0.1in}
\begin{IEEEbiography}[{\includegraphics[width=1in,height=1.25in,clip,keepaspectratio]{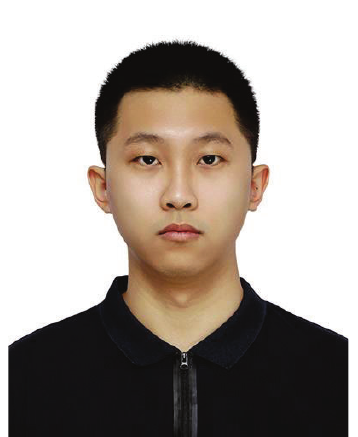}}]{Hanwen Zhang}
received the B.E. degree in Material Science and Engineering with a minor in Robotics from the University of Science and Technology Beijing, Beijing, China, in 2022. He is currently working towards the M.Sc. degree in Robotics at the Department of Mechanical Engineering, National University of Singapore (NUS), Singapore. 

Since August 2023, he has been with the Department of Mechanical Engineering, NUS, Singapore. His research interests include intelligent robotic systems, machine learning, and motion planning.
\end{IEEEbiography}

\begin{IEEEbiography}[{\includegraphics[width=1in,height=1.25in,clip,keepaspectratio]{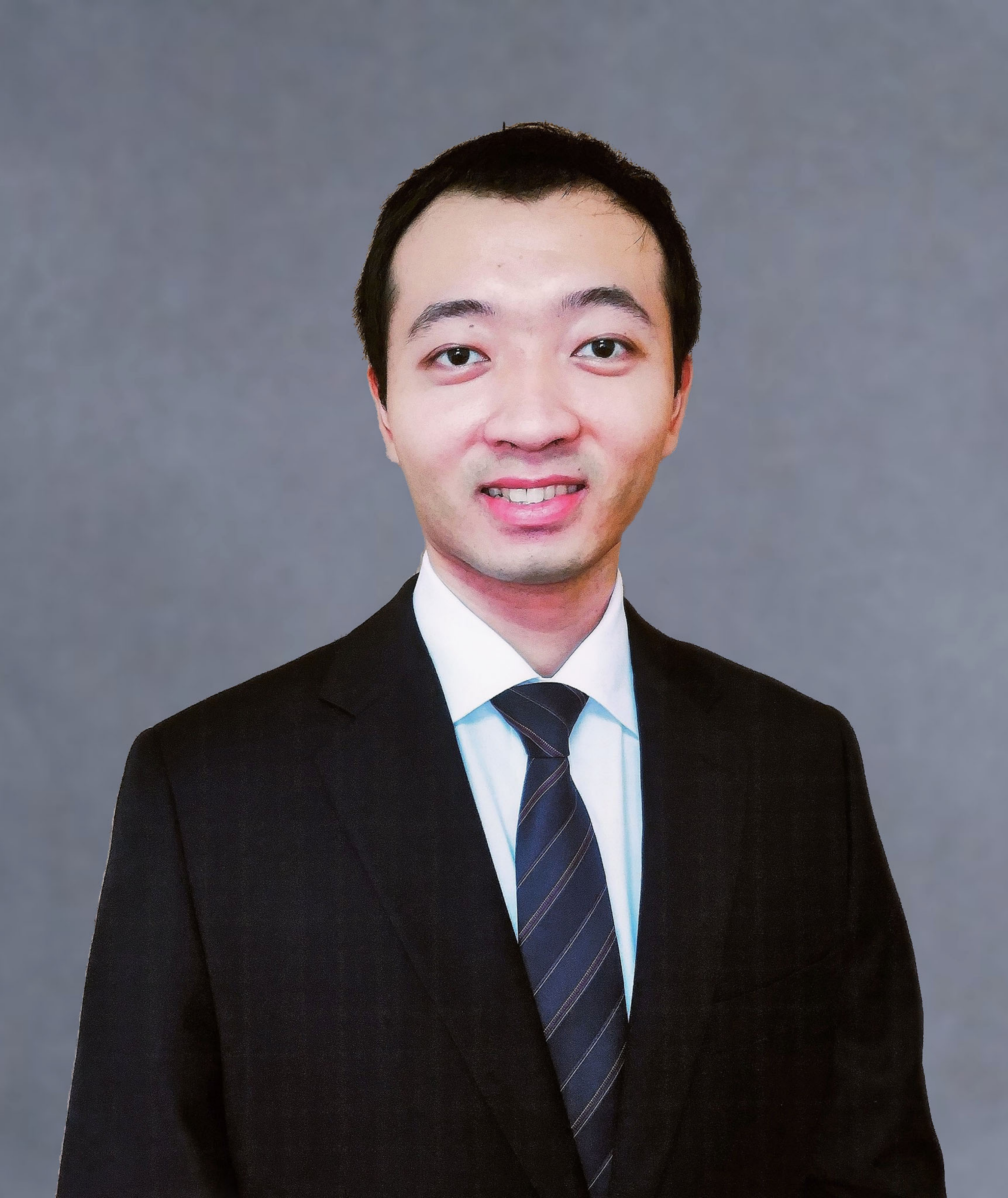}}]{Haozhe Wang} (Graduate Student Member, IEEE)
received the B.E. (Hons.) degree in Electrical Engineering (Distinction) with a specialization in Internet of Things from the National University of Singapore (NUS) in 2021. He is currently working towards the Ph.D. degree in the Integrative Sciences and Engineering Programme (ISEP), NUS Graduate School, Singapore.

Since August 2021, he has been with the ISEP and NUS Advanced Robotics Centre. His current research interests include robotic grasping and manipulation, computer vision, and learning from demonstrations.
\end{IEEEbiography}

\vspace{-0.1in}
\begin{IEEEbiography}[{\includegraphics[width=1in,height=1.25in,clip,keepaspectratio]{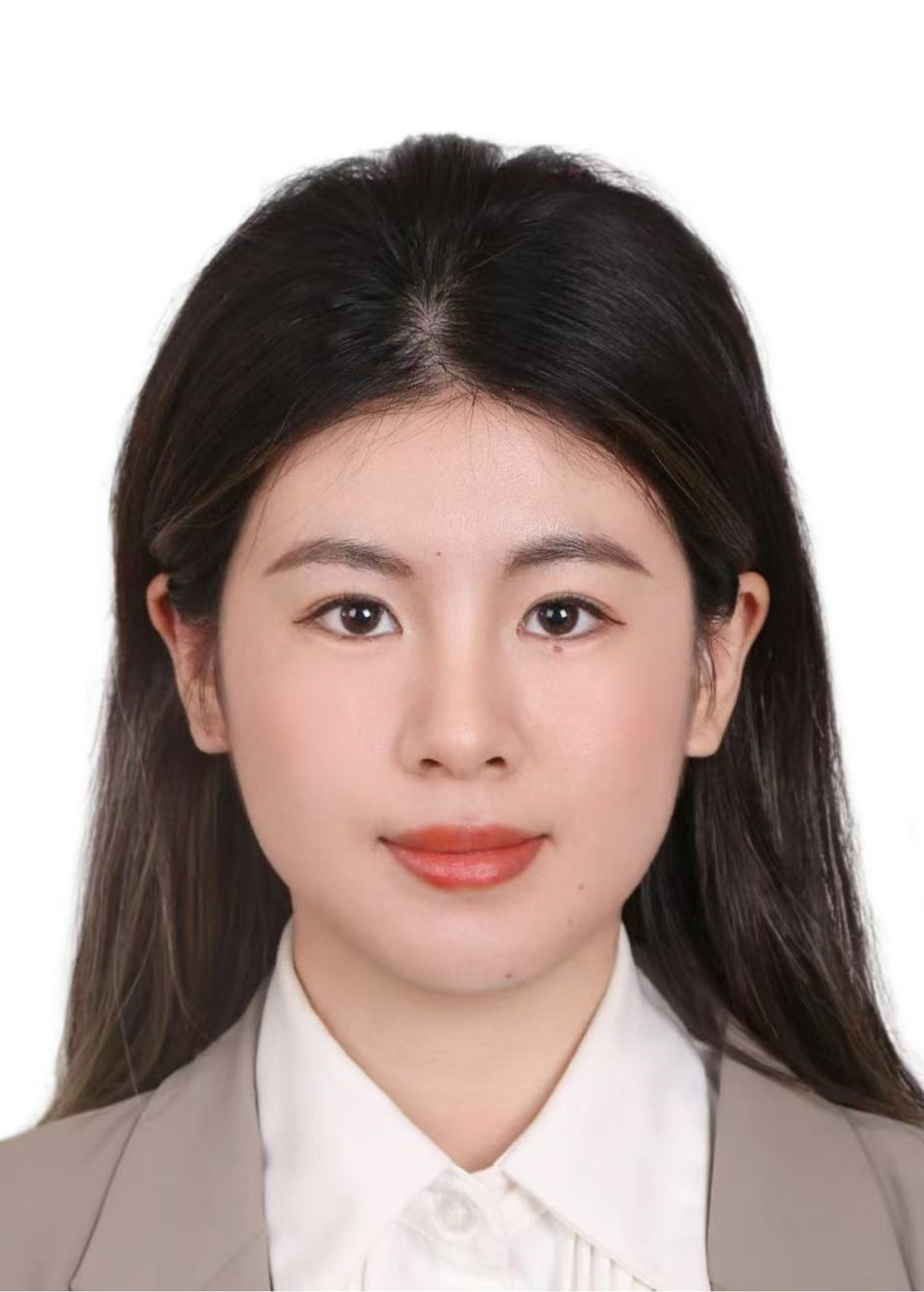}}]{Huishi Huang} (Graduate Student Member, IEEE)
received the M.E. degree in Physics and Engineering in Medicine from University College London, UK, in 2022. She is currently working towards the Ph.D. degree in Mechanical Engineering at the National University of Singapore (NUS), Singapore. 

Since August 2023, she has been with the Department of Mechanical Engineering, NUS, and the Institute of High Performance Computing, A*STAR, Singapore. Her research interests include bio-inspired soft manipulators, machine learning, and mechatronics. 
\end{IEEEbiography}

\begin{IEEEbiography}[{\includegraphics[width=1in,height=1.25in,clip,keepaspectratio]{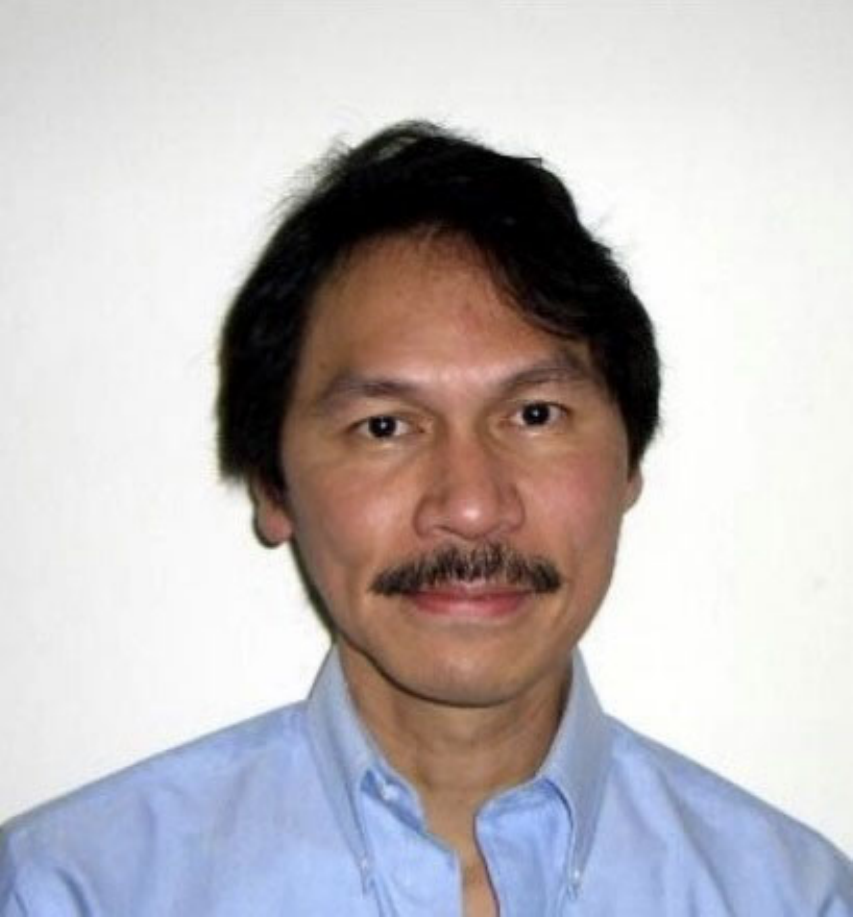}}]{Marcelo H. Ang Jr.} (Senior Member, IEEE)
received the B.Sc. degree (\textit{Cum Laude}) in Mechanical Engineering and Industrial Management Engineering from the De La Salle University, Manila, Philippines, in 1981, the M.Sc. degree in Mechanical Engineering from the University of Hawaii at Manoa, Honolulu, Hawaii, in 1985, and the M.Sc. and Ph.D. degrees in Electrical Engineering from the University of Rochester, Rochester, NY, USA, in 1986 and 1988, respectively. His work experience includes heading the Technical Training Division of Intel’s Assembly and Test Facility in the Philippines, research positions with the East-West Center, Honolulu, and the Massachusetts Institute of Technology, Cambridge, MA, USA, and a Faculty Position as an Assistant Professor of electrical engineering with the University of Rochester. In 1989, Dr. Ang joined the Department of Mechanical Engineering at the National University of Singapore, Singapore, where he is currently a Professor. He is also the Director of the Advanced Robotics Centre. His research interests include robotics, mechatronics, and the application of intelligent systems methodologies. He teaches both graduate and undergraduate levels in the following areas: robotics, creativity and innovation, and engineering mathematics. He is also active in consulting work in robotics and intelligent systems. In addition to academic and research activities, he is actively involved in the Singapore Robotic Games as its founding Chairman and the World Robot Olympiad as a Member of the Advisory Council.
\end{IEEEbiography}

\end{document}